\documentclass[letterpaper]{article} 

\let\savedaddcontentsline\addcontentsline

\usepackage{aaai2027}  
\usepackage[hyphens]{url}  
\usepackage{graphicx} 
\usepackage{natbib}  
\usepackage{caption} 
\usepackage{algorithm}
\usepackage{algorithmic}

\usepackage{newfloat}
\usepackage{amsfonts}
\usepackage{listings}
\DeclareCaptionStyle{ruled}{labelfont=normalfont,labelsep=colon,strut=off} 
\floatstyle{ruled}
\newfloat{listing}{tb}{lst}{}
\floatname{listing}{Listing}

\usepackage{booktabs}
\usepackage{cuted}

\usepackage{pifont}
\usepackage{xcolor}
\usepackage{colortbl}
\usepackage{makecell}
\usepackage{adjustbox}

\usepackage{algorithm}
\usepackage{algorithmic}
\usepackage{amsmath}

\usepackage{titletoc}
\usepackage{multirow}

\definecolor{checkgreen}{RGB}{35,145,45}
\definecolor{crossred}{RGB}{190,35,30}

\newcommand{\cmark}{\textcolor{checkgreen}{\ding{51}}}
\newcommand{\xmark}{\textcolor{crossred}{\ding{55}}}

\newcommand{\eg}{\textit{e.g.}}

\title{MBA: Multimodal Benchmark and Agents for Real-World Business Ideation}

\author{
    Hojun Choi\textsuperscript{\rm 1},
    Jaeyo Shin\textsuperscript{\rm 1},
    Suin Lee\textsuperscript{\rm 1},
    Hyunjung Shim\textsuperscript{\rm 1}\corresponding
}

\affiliations{
    \textsuperscript{\rm 1}KAIST AI, Republic of Korea\\
    \{hchoi256, jaeyo\_shin, suinlee, kateshim\}@kaist.ac.kr\\
    \corresponding Corresponding author
}

\begin{document}

\maketitle

\begin{strip}
  \centering
  \includegraphics[width=\textwidth]{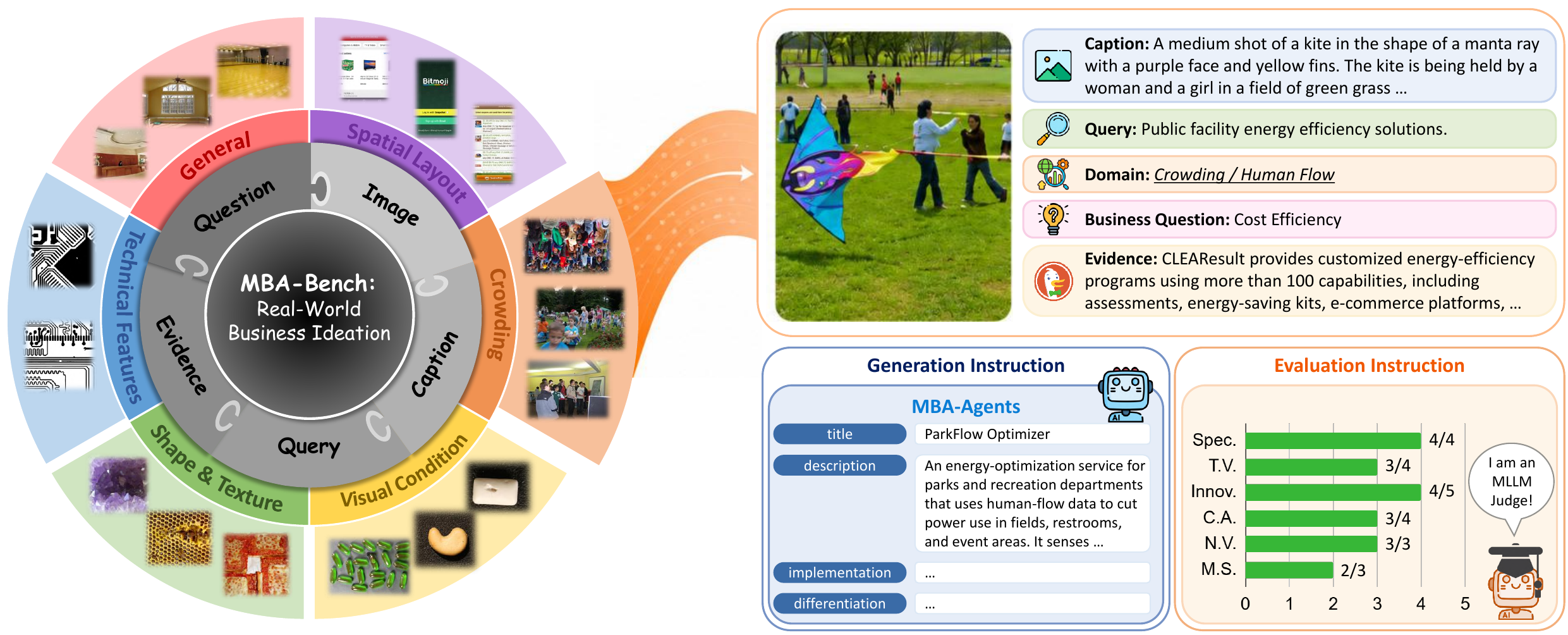}
  \captionof{figure}{
  We introduce MBA-Bench, the first benchmark for training and evaluating multimodal business ideation, comprising 30K samples across six domains, each with a unified question prompt integrating an image, caption, domain, query, business question, and market evidence, evaluated across six metrics (Table~\ref{tab:evaluation_rewards}). Building on this benchmark, we propose two MBA agents that generate both creative and viable business ideas from multimodal inputs, each evaluated via multimodal LLM-as-a-Judge.}
  \label{fig:teaser}
\end{strip}
\begin{abstract}
Agentic systems powered by large language models (LLMs) have opened new opportunities for business ideation. Yet existing approaches remain confined to a text-only paradigm, despite the inherently multimodal nature of real-world contexts. We thus introduce MBA-Bench, the first multimodal benchmark for training and evaluating business ideation agents, comprising 30K samples across six domains, each domain characterized by distinct visual cues not fully conveyed by text alone. Concretely, we automatically caption images and employ GPT-4o to generate five reference ideas for each of three business questions through retrieval query generation, market evidence retrieval, and evidence-augmented synthesis. Following prior work, we evaluate agents across six business-oriented criteria using MLLM-as-a-Judge. To consider settings where criteria are hidden or disclosed, we present MBA-b and MBA-k for blind and known, respectively. We train both with two novel reward objectives---creativity and feasibility---while MBA-k further optimizes the six disclosed criteria for eight in total. Both are trained via LoRA-based supervised fine-tuning followed by group relative policy optimization with these setting-specific rewards. For extensive experiments on MBA-Bench, we set up two baselines accommodating either captions only or multimodal inputs, with the latter nearing closed-source performance on several metrics. MBA-b and MBA-k outperform caption baselines by 63.9\% and 77.1\%, and multimodal baselines by 25.6\% and 35.8\%, respectively.
\end{abstract}

\begin{links}
    \link{Code}{https://github.com/hchoi256/MBA}
    \link{Datasets}{https://huggingface.co/hchoi256/mba}
\end{links}

\section{Introduction}
Agentic AI augments large language models (LLMs) with autonomous reasoning, planning, and decision-making capabilities. It has proven effective at narrowing large candidate spaces, from diagnosis~\cite{mdagent} to fraud detection~\cite{fraud}. Entrepreneurship poses the same challenge: venture capital firms evaluate roughly 100 deals per completed investment, each requiring months of due diligence~\cite{gompers2020vc}. This bottleneck motivates automated support for iterative idea generation and screening.


Toward this end, early approaches rely on zero-shot LLM prompting for ideation, with human experts manually evaluating the resulting business ideas. More recent works~\cite{agentideate,mk2} automate this process by generating multiple ideas from human-curated patent documents and evaluating them across six business-oriented dimensions using LLM-as-a-Judge~\cite{llmasjudge}. Despite this progress, existing methods remain fundamentally bound to a ``text-in, text-out'' paradigm grounded in patents. That is, they are applicable only to scenarios in which the underlying information is both inherently textual and drawn from hard-to-collect patents. In real-world settings, where ideas can instead be derived from readily available multimodal sources, this assumption is neither realistic nor practical.

\begin{figure}[!t]
  \centering
  \includegraphics[width=\linewidth]{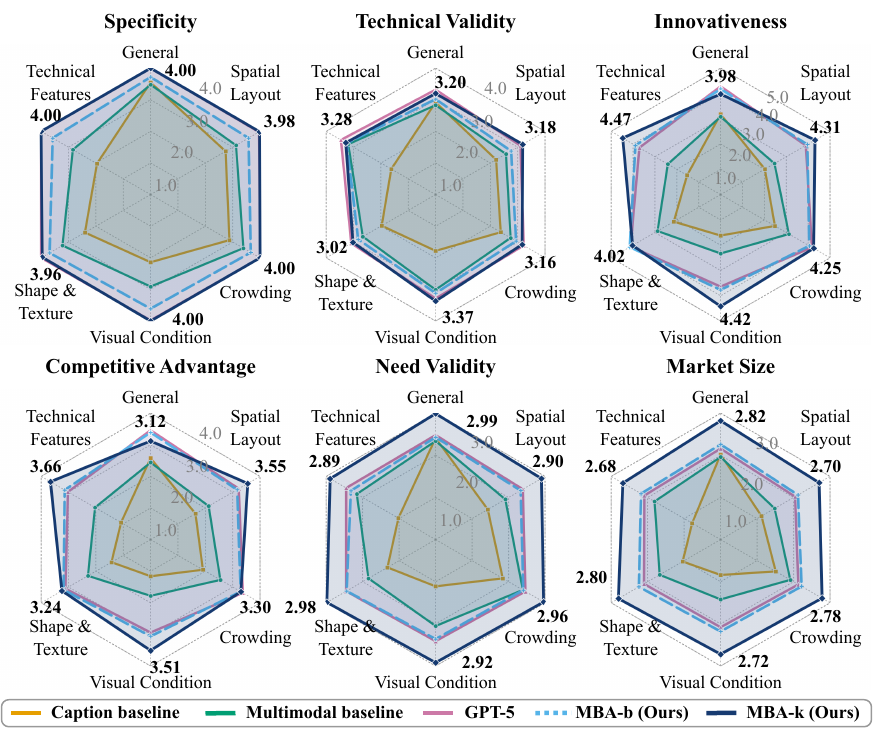}
  \caption{Caption baselines struggle in image-specific domains, while multimodal baselines still lag behind closed-source MLLMs. Our models remain competitive with such MLLMs, opening paths for real-world business ideation.}
  \label{fig:f2}
\end{figure}

We hypothesize that such visual detail is pivotal to distinctive ideation. Put differently, images contain unique information that text cannot fully capture~\cite{multimodallimit}. A natural way to test this premise is to compare each image against its caption for visual fidelity. Specifically, one could attach a strong image captioner~\cite{captioner} to each image and feed the resulting caption into the existing pipeline. In Figure~\ref{fig:f6}, captions may omit details in overly complex scenes (\eg, ethnicity, signage placement), or name a detail too complex to verbalize (\eg, swirling), precluding the opportunities such details could inspire. In practice, the caption baseline trails across all six business-oriented metrics (Figure~\ref{fig:f2})---text alone cannot sustain competitive ideation.

This modality gap motivates revisiting business ideation as a multimodal task. Indeed, multimodal inputs alone outperform their caption-based counterpart by a large margin, even approaching closed-source performance on viability (Figure~\ref{fig:f2}). Yet this strong baseline still falls short on creativity, due to its heavy reliance on zero-shot MLLMs that tend to produce conventional, homogeneous ideas~\cite{jiang} (Figure~\ref{fig:f2}). These cliched outputs hold little commercial value once already on the market. This direction thus remains underexplored, with no dedicated benchmarks or models.

We introduce \textbf{MBA-Bench}, a multimodal benchmark for training and evaluating business ideation agents. As illustrated in Figure~\ref{fig:teaser}, it is designed to reflect real-world multimodal environments through six domains, each capturing visual characteristics not fully represented in text: \textit{General}, \textit{Spatial Layout}, \textit{Crowding}, \textit{Visual Condition}, \textit{Shape \& Texture}, and \textit{Technical Features}. Across these domains, we select 2K images using domain-specific relevance scores derived from dataset annotations (Table~\ref{tab:dataset_statistics}). Each selected image is paired with an automatically generated caption and three high-level questions---cost, technology, and user experience. We further employ GPT-4o~\cite{gpt4o} to implement a three-stage business ideation protocol comprising retrieval-query extraction, market-evidence retrieval through the DuckDuckGo search engine, and evidence-augmented generation. This protocol produces five reference ideas for each question. For evaluation, we follow prior work~\cite{pbig-data} by employing an LLM or MLLM as a judge~\cite{mllmasjudge} to automatically assess generated ideas, thereby reducing manual evaluation costs. The underlying rubric comprises six comprehensive, business-oriented evaluation dimensions introduced in PBIG~\cite{pbig} (Table~\ref{tab:evaluation_rewards}).

Beyond naive multimodal prompting, we also propose two task-specialized agents, \textbf{MBA-b} and \textbf{MBA-k}, which generate both novel and viable business ideas from multimodal contexts. Each targets one of two practical deployment settings, depending on whether the evaluation system is \textbf{b}lind or \textbf{k}nown. Where such a rubric is unknown, MBA-b instead optimizes two task-general objectives: \textit{creativity} (novelty relative to reference ideas) and \textit{feasibility} (market relevance and factuality). MBA-k further leverages the same six disclosed metrics, for eight in total. Each reward is then computed by a separate judge model for training. \textit{Feasibility} is the exception: since judges are themselves prone to hallucinating real-world facts, we instead measure it against \textbf{MBA-Library}, our web-sourced \textit{Feasibility}-grounding resource. Both agents follow a common two-stage training protocol: we first adapt an open-source MLLM~\cite{qwen} using LoRA~\cite{lora} through supervised fine-tuning (SFT) and then apply group relative policy optimization (GRPO)~\cite{deepseek} with setting-specific rewards for within-group ranking.

On MBA-Bench, MBA-b outperforms the caption-based and multimodal baselines by 63.9\% and 25.6\%, respectively. MBA-k likewise outperforms them by 77.1\% and 35.8\%, while remaining competitive with closed-source MLLMs.
\begin{itemize}
    \item We present \textbf{MBA-Bench}, the first multimodal benchmark for training and evaluating real-world business ideation agents, comprising 30K multimodal samples across six domains and eight business-oriented dimensions.
    \item We propose \textbf{MBA-b} and \textbf{MBA-k}, two business ideation agents for blind and known evaluation settings, trained with SFT and GRPO under setting-specific objectives.
    \item Both variants outperform caption-based and multimodal baselines on the MBA-Bench test set, while MBA-k remains competitive with closed-source MLLMs.
\end{itemize}


\section{Related Work}
\subsection{Agentic AI for Business Ideation}
Early agentic AI frameworks extend LLMs with planning and tool use for multi-step reasoning~\cite{react,toolformer}. These text-based capabilities have since extended to entrepreneurship, where AI systems must generate ideas grounded in market needs and originality. Recent work has turned to patent documents as a rich source of technologies intended for commercialization. Concretely, PBIG~\cite{pbig} introduces six business-oriented dimensions for judging patent-derived product ideas. Under these criteria, Agent Ideate~\cite{agentideate} decomposes ideation across specialized, tool-augmented agents, while PBIG-Data~\cite{pbig-data} collects and analyzes expert scores on candidate ideas. By contrast, MK2~\cite{mk2} forgoes an agentic pipeline entirely, instead iteratively refining a single prompt guided by a pairwise judge. Yet text alone cannot fully capture the visual detail of real-world contexts, and no benchmark or agent has tackled this multimodal setting. In this work, we introduce the first such benchmark and agents for real-world, multimodal business ideation.

\subsection{Reinforcement Learning (RL) Post-training}
RL has become a central post-training paradigm for LLMs, first for alignment~\cite{ouyang2022training} and more recently for reasoning. For alignment, proximal policy optimization~\cite{schulman2017proximal} and direct preference optimization~\cite{rafailov2023direct} either pair a reward model with a value network, or discard both for pairwise preference data. For reasoning, group relative policy optimization (GRPO)~\cite{deepseek} keeps a scalar reward but drops the value network, estimating advantages from relative scores within a sampled group. This simplicity has driven GRPO's extension to MLLMs across both single-answer and open-ended settings. Concretely, Visual-RFT~\cite{visualrft} applies rule-based rewards to visual perception tasks with a definite answer, whereas Debate-as-Reward~\cite{debateasreward} trains a multi-agent judge to reward scientific idea generation, where no single answer is correct. Business ideation similarly admits no single correct answer: a proposal should be both creative and feasible, neither fully verifiable. In this work, we optimize an MLLM with GRPO under judge-based rewards~\cite{mllmasjudge} tailored to these two objectives.


\section{Methodology}
We recast business ideation as a multimodal task grounded in real-world contexts, with a dedicated benchmark and two task-specific agents. The underlying motivation is to uncover hints lost in cross-modal translation or either modality alone.

\begin{figure*}[!t]
  \centering
  \includegraphics[width=\linewidth]{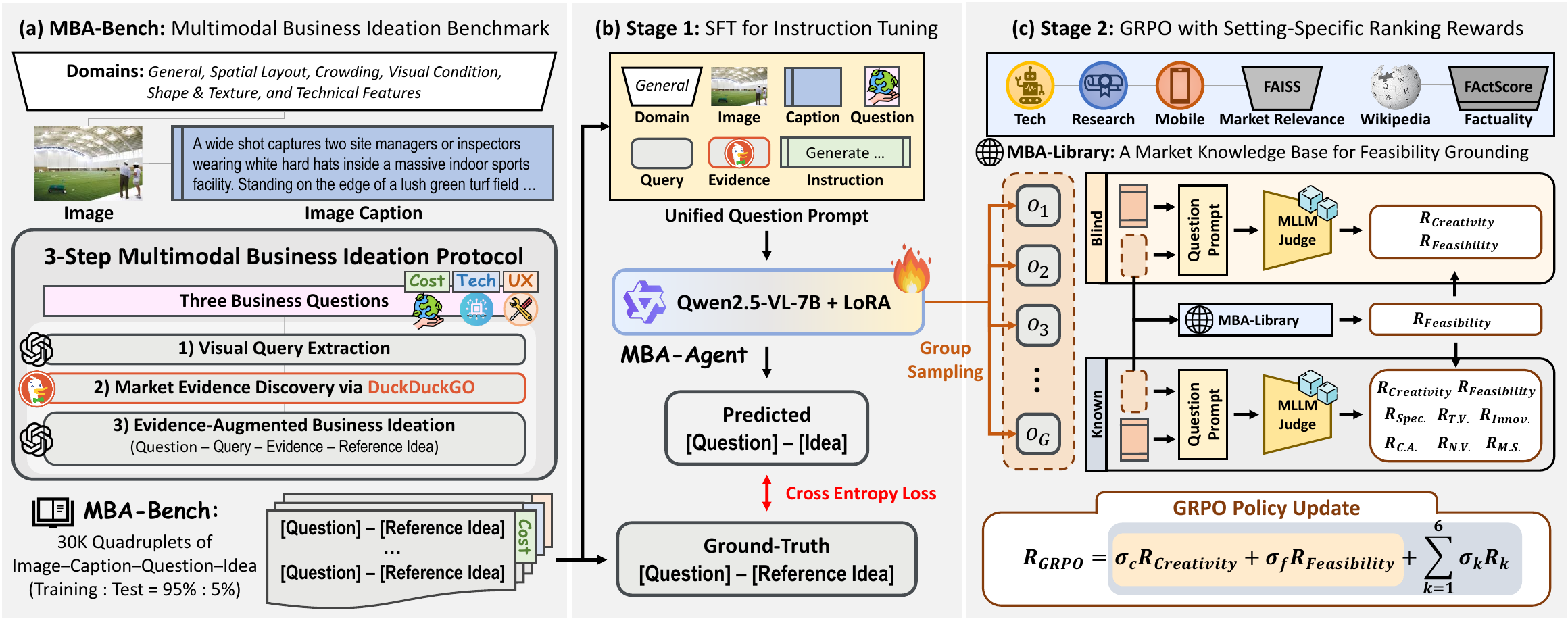}
  \caption{\textbf{Overview of MBA.} \textbf{(a)} We present MBA-Bench, comprising 30K image--caption--question--idea quadruplets across six domains curated via a three-step protocol using GPT-4o and DuckDuckGo. \textbf{(b)} We then instruction-tune the base MLLM on question--idea pairs via SFT. \textbf{(c)} We further leverage GRPO to refine the SFT checkpoint with setting-specific ranking rewards---\textit{creativity} and \textit{feasibility} for the blind agent MBA-b, plus six disclosed dimensions for the known agent MBA-k.}
  \label{fig:main}
\end{figure*}

We introduce \textbf{MBA-Bench}, a multimodal benchmark for training and evaluation in business ideation (Figure~\ref{fig:main}-a). We sample 2K images across six domains in predefined proportions based on annotation-driven scores. Each is then paired with a caption from a captioner~\cite{captioner}. For each pair, we formulate three business questions and generate five reference ideas per question using GPT-4o through three stages---visual query extraction, market evidence retrieval via the DuckDuckGo API, and evidence-augmented ideation---yielding 30K image--caption--question--idea quadruplets. For evaluation, we use an MLLM-as-a-Judge~\cite{mllmasjudge} to assess ideas across six business-oriented dimensions~\cite{pbig}.

Furthermore, we propose two dedicated agents, \textbf{MBA-b} and \textbf{MBA-k}, for blind and known evaluation settings, respectively. Both models share two reward objectives---\textit{creativity} and \textit{feasibility}---while MBA-k further incorporates rewards for the six disclosed criteria. In both cases, training follows the same two-stage pipeline: LoRA-based~\cite{lora} supervised fine-tuning (SFT) of an open-source MLLM~\cite{qwen} (Figure~\ref{fig:main}-b) and group relative policy optimization (GRPO) with setting-specific rewards (Figure~\ref{fig:main}-c).

\subsection{MBA-Bench: Multimodal Business Ideation Benchmark}
\label{method:s1}
We introduce \textbf{MBA-Bench}, the first benchmark for training and evaluation on multimodal business ideation. We begin by elaborating on our data curation pipeline across six domains, each highlighting distinct characteristics. We then design a retrieval-augmented ideation protocol grounded in market information. Finally, we present the evaluation framework and six metrics used to assess agents on this benchmark.

\subsubsection{Domain-Specific Multimodal Data Curation}
Real-world images provide an untapped opportunity for extending business ideation beyond the prevailing text-dependent paradigm. Since text alone cannot capture the full range of visual detail, we curate image--caption pairs across six domains, from visual content that is easily verbalized to that which is not.

As detailed in Table~\ref{tab:dataset_statistics}, \textit{General} comprises ADE20K everyday scenes~\cite{ade20k}, effectively represented in both visual and textual form. The remaining five domains target visual semantics hard to verbalize: \textit{Spatial Layout} consists of RICO screenshots~\cite{rico} with densely arranged mobile interface components; \textit{Crowding} comprises COCO images~\cite{coco} with at least ten annotated people; \textit{Visual Condition} uses VisA anomaly images~\cite{visa} depicting subtle surface defects; \textit{Shape \& Texture} contains close-up material surfaces from DTD~\cite{dtd} with fine-grained patterns; and \textit{Technical Features} consists of circuit-board images from DeepPCB~\cite{deeppcb} with intricate structural details. Together, these domains span varying degrees of verbalizability for comparative analysis.

For each domain, we select representative samples by ranking images on dataset annotations. The criteria are object diversity for \textit{General}; component count and text density for \textit{Spatial Layout}; person count and object diversity for \textit{Crowding}; a fixed ratio favoring anomalous cases for \textit{Visual Condition}; uniform coverage of texture attributes for \textit{Shape \& Texture}; and prioritization of defect-annotated images for \textit{Technical Features}. Through image captioning, this process finally yields 2K image--caption pairs across six domains.

\subsubsection{Retrieval-Augmented Ideation Protocol}
Given each image–caption pair, we develop a business ideation protocol that defines the task on which agents are trained and evaluated. At its core, the protocol requires generating reference ideas grounded in market evidence rather than model speculation; these ideas serve as supervision only during training. Accordingly, prior work~\cite{agentideate} leverages retrieval-augmented generation (RAG) to ground ideas in market information via web search APIs like DuckDuckGo.

Building on this approach, we design a three-stage RAG pipeline for multimodal business ideation (Figure~\ref{fig:main}-a). First, we employ GPT-4o to derive retrieval queries from salient visual observations in each image (\eg, ``a long customer queue at a service counter''). These queries are then submitted to the DuckDuckGo API to retrieve relevant market evidence, such as existing solutions and wait-time statistics. Since a single piece of evidence can motivate ideation across diverse angles, we structure this diversity into three broad, well-established question types---cost efficiency, technology, and user experience~\cite{question}. For each question, we construct a unified prompt integrating the four previously obtained elements---the image--caption pair, domain, query, and evidence---from which GPT-4o generates five reference ideas. Following prior work~\cite{agentideate}, each idea comprises four fields---\textit{title}, \textit{description}, \textit{implementation}, and \textit{differentiation}---thereby ensuring sufficiently detailed and well-organized ideas (Figure~\ref{fig:f4}). Altogether, this process yields 30K image--caption--question--idea quadruplets from 2K samples (Table~\ref{tab:dataset_statistics}), laying the groundwork for training and evaluating business ideation agents at scale.

\begin{table}[!t]
\centering

\begin{adjustbox}{max width=\columnwidth,center}
\begin{tabular}{@{}llcccc@{}}
    \toprule
    Domain
    & Source Dataset
    & Images
    & Captions
    & \makecell{Business\\Questions}
    & \makecell{Reference\\Ideas} \\
    \midrule

    \textit{General}
    & ADE20K~\citeyearpar{ade20k}
    & 500
    & 500
    & 1,500
    & 7,500 \\

    \textit{Spatial Layout}
    & RICO~\citeyearpar{rico}
    & 350
    & 350
    & 1,050
    & 5,250 \\

    \textit{Crowding}
    & COCO~\citeyearpar{coco}
    & 350
    & 350
    & 1,050
    & 5,250 \\

    \textit{Visual Condition}
    & VisA~\citeyearpar{visa}
    & 350
    & 350
    & 1,050
    & 5,250 \\

    \textit{Shape \& Texture}
    & DTD~\citeyearpar{dtd}
    & 350
    & 350
    & 1,050
    & 5,250 \\

    \textit{Technical Features}
    & DeepPCB~\citeyearpar{deeppcb}
    & 100
    & 100
    & 300
    & 1,500 \\
    \midrule

    -
    & -
    & 2,000
    & 2,000
    & 6,000
    & 30,000 \\
    \bottomrule
\end{tabular}
\end{adjustbox}

\caption{Dataset statistics of MBA-Bench by domain.}
\label{tab:dataset_statistics}
\end{table}

\subsubsection{Business Idea Evaluation via MLLM-as-a-Judge}
Manual evaluation by human annotators is inherently subjective and costly at scale. Recent work~\cite{pbig-data} has thus adopted the LLM-as-a-Judge paradigm~\cite{llmasjudge} for automatically evaluating generated business ideas.

To account for the multimodal nature of real-world business ideation, we extend this paradigm to an MLLM-as-a-Judge~\cite{mllmasjudge}. On MBA-Bench, an agent is first asked to generate a business idea for each unified question prompt. The generated idea is then evaluated by a strong 78B-parameter MLLM~\cite{internvl}, using the same prompt with the evaluation instruction. The judge assesses each idea along six business-oriented dimensions~\cite{pbig} on their respective scoring scales (Table~\ref{tab:evaluation_rewards}). This evaluation protocol enables automated, scalable assessment of business ideas within their multimodal and market context.

\subsection{MBA Agents: Multimodal Business Ideation Agents}
\label{method:s2}
Multimodal inputs enable zero-shot MLLMs to discover business opportunities from visual contexts (Figure~\ref{fig:f2}). Despite strong generalization, they lag behind closed-source MLLMs in \textit{innovativeness} and \textit{competitive advantage}. The cause lies in general-purpose MLLMs' tendency to produce conventional outputs~\cite{jiang,huang}. This limitation motivates a dedicated model for the task.

Designing such a model requires addressing a key uncertainty: the real-world evaluation criteria are not always known in advance. If undisclosed, an agent must be trained without ever accessing them. We therefore consider two real-world settings---\textbf{b}lind and \textbf{k}nown---and propose a dedicated agent for each, \textbf{MBA-b} and \textbf{MBA-k}, respectively. Concretely, we first define two task-general objectives applicable regardless of setting: \textit{creativity} (novelty relative to reference ideas) and \textit{feasibility} (market relevance and factual consistency). MBA-b is then trained with this dual-objective reward alone, while MBA-k additionally optimizes the six evaluation metrics (Table~\ref{tab:evaluation_rewards}). Each reward is computed by an MLLM judge, except \textit{feasibility}, measured against \textbf{MBA-Library}---our web-sourced resource. Finally, our training pipeline involves two stages on an open-source MLLM~\cite{qwen}: LoRA-based supervised fine-tuning (SFT)~\cite{lora}, followed by group relative policy optimization (GRPO)~\cite{deepseek} with setting-specific rewards for within-group ranking.

\subsubsection{Quantifying Reward Objectives}
To quantify each objective as a reward within its range (Table~\ref{tab:evaluation_rewards}), we employ separate MLLM judges for training and evaluation, avoiding model bias from reusing the same judge. Under this mechanism, the six known criteria are scored using the same instruction as prior work~\cite{pbig}, while \textit{creativity} measures how novel the idea is relative to its five reference ideas.

Ironically, MLLMs are inherently prone to generating irrelevant or hallucinated content~\cite{mllmhallu}. This risk may be milder for subjective dimensions such as \textit{innovativeness}, but can be acute wherever real-world grounding is required. \textit{Feasibility} is such a case: both its market relevance and factuality must align with reality. Beyond model-internal knowledge, we design \textbf{MBA-Library}---an enormous web-sourced resource for \textit{feasibility} grounding---spanning mobile applications~\cite{mobilerec}, scientific literature~\cite{openalex}, structured entities~\cite{wikidata}, and Wikipedia. This database powers two scoring modules: one for market relevance and one for factuality. Concretely, we embed each idea and query the library with FAISS~\cite{faiss} to retrieve its top-$k$ nearest records by cosine similarity; market relevance is then scored as the average similarity to these records. Factuality instead follows FActScore~\cite{factscore}: an LLM~\cite{llama} decomposes each idea into atomic facts, which a retriever~\cite{gtr} matches to relevant Wikipedia passages for verification. Both scores are then normalized to a $[0,1]$ range. In short, these rewards support both training and evaluation for multimodal business ideation.

\begin{table}[!t]
\centering

\begin{adjustbox}{max width=\columnwidth,center}
\begin{tabular}{@{}llcccc@{}}
    \toprule
    Dimension
    & Focus
    & Range
    & \multicolumn{2}{c}{Train}
    & Test \\
    \cmidrule(lr){4-5}
    &
    &
    \textnormal{(Cutoff)}
    & Blind
    & Known
    & \\
    \midrule

    \textit{Creativity}$^\dagger$
    & Novelty beyond reference ideas
    & 0--1
    & \cmark
    & \cmark
    & \xmark \\

    \textit{Feasibility}
    & Market relevance \& factuality
    & 0--1
    & \cmark
    & \cmark
    & \xmark \\
    \midrule

    \textit{Specificity}$^\dagger$
    & Clarity of idea
    & 1--4 ($>$ 2)
    & \xmark
    & \cmark
    & \cmark \\

    \textit{Technical Validity}$^\dagger$
    & Feasibility
    & 1--4 ($>$ 1)
    & \xmark
    & \cmark
    & \cmark \\

    \textit{Innovativeness}$^\dagger$
    & Novelty
    & 1--5
    & \xmark
    & \cmark
    & \cmark \\

    \textit{Competitive Advantage}$^\dagger$
    & Differentiation
    & 1--4
    & \xmark
    & \cmark
    & \cmark \\

    \textit{Need Validity}$^\dagger$
    & User need
    & 0--3
    & \xmark
    & \cmark
    & \cmark \\

    \textit{Market Size}$^\dagger$
    & Adoption scale
    & 0--3
    & \xmark
    & \cmark
    & \cmark \\
    \bottomrule
\end{tabular}
\end{adjustbox}

\caption{Business-oriented scoring dimensions by setting. $\dagger$ denotes the use of MLLM-as-a-Judge~\cite{mllmasjudge}.}
\label{tab:evaluation_rewards}
\end{table}

\subsubsection{SFT: Question--Reference Idea Instruction Tuning}
We perform SFT to instruction-tune an open-source MLLM~\cite{qwen} with LoRA~\cite{lora} to generate a business idea given a question. On MBA-Bench, each of the five reference ideas per question serves as an independent target response, yielding 28.5K training instances in total. We optimize the next-token prediction loss over the target tokens:
\begin{equation}
\mathcal{L}_{\mathrm{SFT}} = -\sum_{t=1}^{T} \log p_{\theta,\phi}\left(y_t \mid x, y_{<t}\right),
\end{equation}
where $x = (v, c, b, q, e)$ denotes the image $v$, its caption $c$, the business question $b$, the retrieval query $q$, and the retrieved market evidence $e$; and $y = (y_1, \ldots, y_T)$ denotes the corresponding reference idea with $T$ tokens (Figure~\ref{fig:main}-b).

\subsubsection{GRPO: Setting-Specific Ranking Reward}
To directly tackle the mode collapse toward conventional ideas observed in MLLMs, we further fine-tune the SFT model with GRPO, using the setting-specific rewards introduced above (Figure~\ref{fig:main}-c). For each unified question prompt, the policy first samples a group of $G$ candidate ideas $\{o_1, \ldots, o_G\}$. Rather than scoring each candidate in isolation, the judge then ranks the group along each objective, with each rank converted into a $[0,1]$ score. This relative scoring avoids the judge-specific scale bias, where distinct ideas often receive identical scores.

As specified in Table~\ref{tab:evaluation_rewards}, the setting-specific reward $r_i$ for candidate $o_i$ combines two objectives for MBA-b, or all eight for MBA-k. Within each group, these rewards are then normalized into an advantage of the $i$-th candidate's relative strength, with group mean $\mu$ and standard deviation $\sigma$:
\begin{equation}
\label{eq:grpo_advantage}
A_i = \frac{r_i - \mu(\{r_j\}_{j=1}^{G})}{\sigma(\{r_j\}_{j=1}^{G}) + \epsilon}.
\end{equation}
Using this advantage, we update the GRPO policy against a frozen reference $\pi_{\mathrm{ref}}$ initialized from the SFT checkpoint:
\begin{equation}
\label{eq:grpo_objective}
\mathcal{L}_{\mathrm{GRPO}} = -\mathbb{E}_i\left[A_i \log \pi_\theta(o_i \mid x)\right] + \beta \, \mathbb{E}_i\left[\mathbb{D}_{\mathrm{KL}}\left(\pi_\theta \,\|\, \pi_{\mathrm{ref}}\right)\right],
\end{equation}
where $\pi_\theta$ is the policy being optimized and $\beta$ controls the strength of KL regularization toward $\pi_{\mathrm{ref}}$. As a result, this group-relative optimization steers MBA-b and MBA-k beyond homogeneous priors toward creative, feasible ideas.


\begin{figure}[!t]
  \centering
  \includegraphics[width=\linewidth]{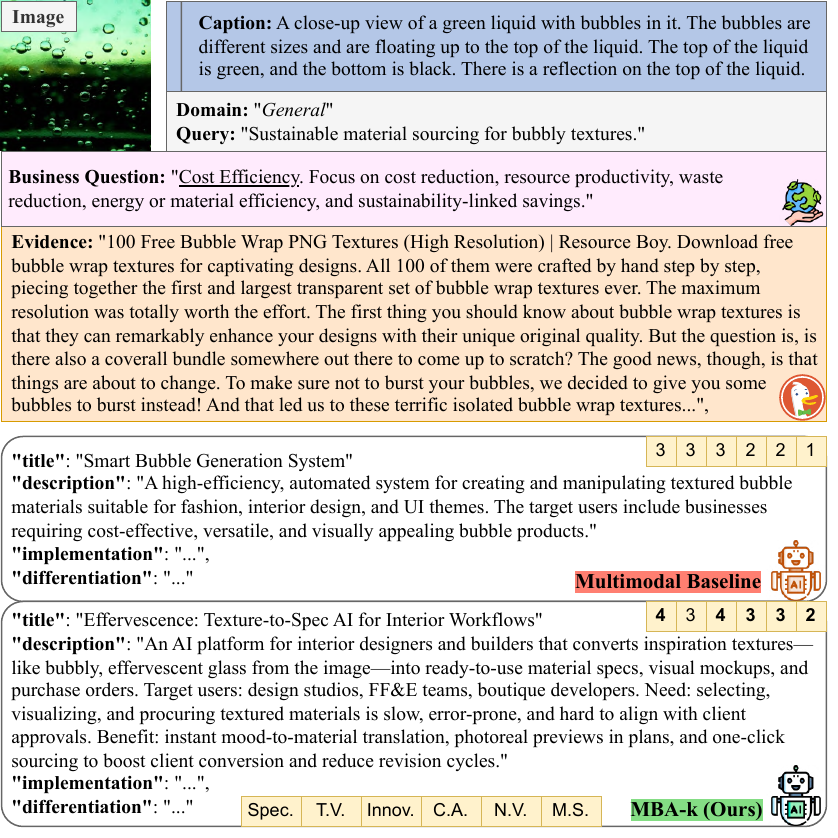}
  \caption{The upper section indicates the unified question prompt: the image--caption pair, domain, query, question, and evidence. The lower section compares a multimodal baseline with our agent across the six metrics in the top-right box.}
  \label{fig:f4}
\end{figure}

\section{Experiments}
We first analyze the statistical characteristics of MBA-Bench across domains and metrics. We then evaluate diverse MLLMs on MBA-Bench, showing that our agents come closest to distinctive ideation. Through ablation studies, we validate our method across modalities. Finally, we examine modality mismatch, revealing visual cues that fail to transfer.

\subsection{Benchmark Analysis}
In this section, we provide a statistical analysis of MBA-Bench spanning all domains, components, and metrics. We also specify the implementation details of our MBA agents.

\subsubsection{Datasets \& Evaluation Metrics}
MBA-Bench comprises 30K image--caption--question--idea quadruplets across six domains, where each domain contains a predefined number of images selected from existing datasets based on dataset annotations (Table~\ref{tab:dataset_statistics}). Each image has its caption, three business questions, and five reference ideas per question, totaling 15 ideas per image. Also, each image--caption pair is accompanied by four supporting elements---domain, query, question, and evidence---constituting the unified question prompt for business ideation (Figures~\ref{fig:main} and~\ref{fig:f4}). These data are split by image into training and test sets at a 95\%--5\% ratio, respectively. Moreover, we follow the PBIG rubric~\cite{pbig} to evaluate ideas (Table~\ref{tab:evaluation_rewards}), but with an MLLM judge.

\subsubsection{Implementation Details}
We use PaliGemma2~\cite{captioner} as a long-context image captioner. We employ GPT-4o to develop MBA-Bench. Our agents are initialized from Qwen2-VL-7B-Instruct using LoRA~\cite{lora} with rank 32, scaling factor 64, and dropout 0.05. For judge-based rewards, we use Qwen2-VL-72B-Instruct~\cite{qwen} during training and InternVL2.5-78B~\cite{internvl} during evaluation. For \textit{feasibility}, a non-judge metric, our MBA-Library leverages FAISS~\cite{faiss} and FActScore~\cite{factscore} to assess the market relevance and factuality of an idea against the large web-sourced dataset. For SFT, we train the LoRA adapters for 2 epochs with a learning rate of $2\times10^{-5}$, a warmup ratio of 0.1, and a global batch size of 32. We then continue training with GRPO for 1 epoch using a learning rate of $1\times10^{-6}$, $G=4$ responses per prompt, a KL coefficient of 0.02, and a batch size of 4. The GRPO weights for Spec., T.V., Innov., C.A., N.V., M.S., \textit{Creativity}, and \textit{Feasibility} are 0.12, 0.12, 0.20, 0.16, 0.12, 0.08, 0.10, and 0.10 in the known case, respectively; the blind case retains only the last two, weighted 0.70 and 0.30. All reported results are based on a single run.

\begin{table}[!t]
\centering

\begin{adjustbox}{max width=\columnwidth,center}
\begin{tabular}{@{}lcccccc@{}}
\toprule
Model
& Spec.
& T.V.
& Innov.
& C.A.
& N.V.
& M.S. \\
\midrule

GPT4o
& 3.51
& 3.04
& 3.04
& 2.79
& 2.27
& 2.05 \\

GPT5-mini
& \textbf{4.00}
& 3.07
& 3.59
& 3.00
& 2.77
& 2.07 \\

GPT5
& \underline{3.99}
& 3.08
& 3.97
& \underline{3.27}
& 2.44
& 2.10 \\

Claude-Sonnet-4.6
& 3.88
& \textbf{3.16}
& 3.78
& 3.24
& 2.58
& 2.16 \\

Gemini-3.5-Flash
& \textbf{4.00}
& 3.05
& \textbf{4.00}
& 3.01
& \textbf{2.97}
& 2.29 \\

Gemini-3.1-pro-preview
& \textbf{4.00}
& 3.04
& \underline{3.98}
& 3.01
& \textbf{2.97}
& \underline{2.35} \\

Gemini-3.6-Flash
& \textbf{4.00}
& 3.03
& \textbf{4.00}
& 3.01
& 2.57
& 2.34 \\

\midrule

LLaVA-OneVision-Qwen2-7B
& 3.39
& 2.99
& 3.17
& 2.91
& 2.33
& 2.15 \\

LLaVA-OneVision-Qwen2-7B
& 3.39
& 2.99
& 3.17
& 2.91
& 2.33
& 2.15 \\

LLaVA-NeXT-Qwen-32B
& 3.28
& 2.97
& 3.19
& 2.91
& 2.22
& 2.07 \\

InternVL2.5-8B
& 3.62
& 3.01
& 3.19
& 2.95
& 2.39
& 2.12 \\

InternVL2.5-26B
& 3.58
& 3.01
& 3.27
& 2.97
& 2.46
& 2.20 \\

Qwen2.5-VL-7B-Instruct
& 3.60
& 3.06
& 3.15
& 2.62
& 2.32
& 1.94 \\

Qwen2.5-VL-32B-Instruct
& 3.50
& 2.99
& 3.23
& 2.96
& 2.21
& 2.08 \\

\rowcolor{yellow!10}
\textbf{MBA-7B-SFT (Ours)}
& 3.68
& 3.09
& 3.22
& 2.57
& 2.31
& 1.99 \\

\rowcolor{cyan!10}
\textbf{MBA-b-7B (Ours)}
& 3.64
& \underline{3.12}
& \underline{3.98}
& 3.19
& 2.38
& 2.20 \\

\rowcolor{cyan!10}
\textbf{MBA-k-7B (Ours)}
& \underline{3.99}
& 3.00
& \textbf{4.00}
& \textbf{3.32}
& \underline{2.94}
& \textbf{2.75} \\

\bottomrule
\end{tabular}
\end{adjustbox}

\caption{
Benchmark results on the MBA-Bench test set, following the scoring
scales in Table~\ref{tab:evaluation_rewards}.
Spec.: \textit{Specificity};
T.V.: \textit{Technical Validity};
Innov.: \textit{Innovativeness};
C.A.: \textit{Competitive Advantage};
N.V.: \textit{Need Validity};
and M.S.: \textit{Market Size}.
}
\label{tab:business_evaluation_results}
\end{table}

\subsection{Main Results}
Table~\ref{tab:business_evaluation_results} compares open-source MLLMs with MBA-b and MBA-k across six metrics, reporting means and standard deviations across images. The standard deviations indicate generally consistent performance across images. For a fair comparison, all models receive the same multimodal inputs and unified question prompt. Instruction-tuned on well-organized GPT-4o-generated reference ideas, MBA-SFT outperforms the baseline on all four feasibility metrics—Spec., T.V., N.V., and M.S. MBA-b further achieves state-of-the-art performance among open-source models on nearly all metrics, particularly in Innov. and C.A. These improvements over the baseline validate our training-time rewards for creative and feasible business ideation. Moreover, MBA-k, a stronger agent trained using all eight rewards, achieves the best performance on both \textit{Creativity}- and \textit{Feasibility}-related metrics among all models. This result shows that our model effectively adapt to the target environment when its evaluation criteria are known---an effective direction for advancing business ideation, even at a compact 7B-parameter scale.

\begin{figure*}[!t]
  \centering
  \includegraphics[width=\linewidth]{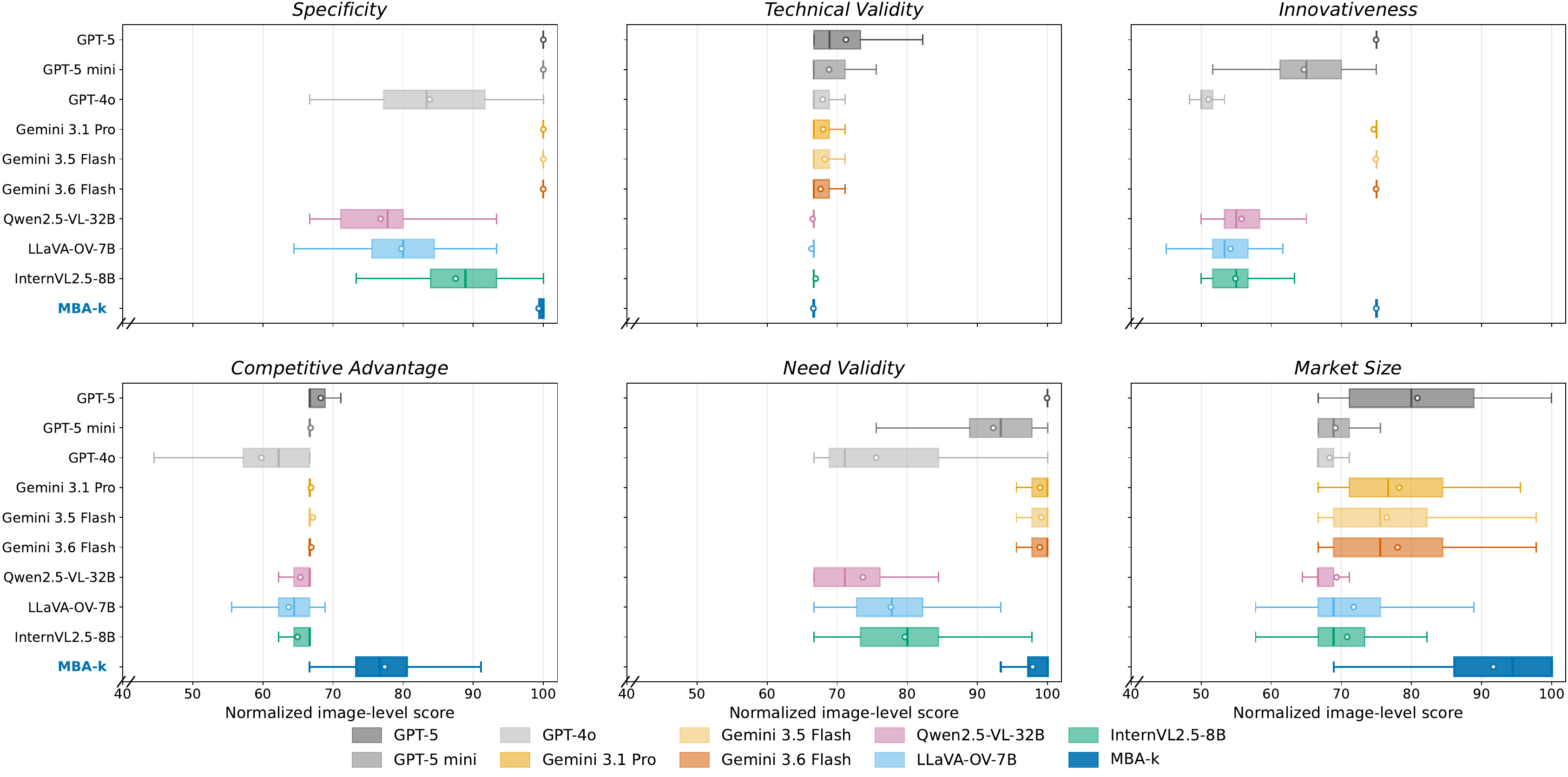}
  \caption{Image-level distributions of six PBIG metrics across ten models. For each image, scores are averaged over 15 responses and normalized to a 0--100 scale. Boxes show the interquartile range, center lines the medians, and circles the means.}
  \label{fig:supp:boxplot}
\end{figure*}

Figure~\ref{fig:supp:boxplot} compares the image-level score distributions of MBA-k with those of nine proprietary and open-source models. For each of the 100 test images, scores are averaged over 15 responses, and the six metrics are normalized to a common 0--100 scale. MBA-k shows the clearest distributional advantages in \emph{Competitive Advantage} and \emph{Market Size}, with substantially higher medians and means than all comparison models. It also remains competitive with the strongest proprietary models in \emph{Innovativeness} and \emph{Need Validity}. In contrast, \emph{Specificity} is tightly concentrated near the upper bound for several models, suggesting a ceiling effect, while MBA-k shows a modest trade-off in \emph{Technical Validity}. Overall, these results highlight MBA-k's strong performance in differentiated and market-oriented business ideation, with clear advantages in dimensions closely tied to business value while remaining broadly competitive across the other metrics.

For human evaluation, prior work~\cite{mk2} offers a useful reference point: under the same six PBIG criteria, a strong GPT-based system ranked first in five of six human-evaluated criteria in natural language processing and four of six in computer science. Table~\ref{tab:business_evaluation_results} demonstrates that MBA-b and MBA-k remain competitive with such systems on the same criteria. This suggests our agents' ideas could similarly earn favorable ratings from human experts. Direct human evaluation of our agents on MBA-Bench remains future work.

\subsection{Ablation Study}
In this section, we analyze our agents across domains, modalities, and internal components. We also validate our training objectives and qualitatively examine cross-modal mismatch.

\subsubsection{Domain-wise Analysis}
In Figure~\ref{fig:f2}, we compare our agents across six domains for each metric. Domain and metric definitions are provided in Tables~\ref{tab:dataset_statistics} and~\ref{tab:evaluation_rewards}. For Spec., MBA-k achieves near-perfect, highly consistent scores across all domains, while MBA-b exhibits a similarly uniform but slightly lower pattern. This uniformity indicates limited sensitivity to domain variation for this metric. N.V. and M.S.---which focus on user needs and adoption scale, respectively---peak in \textit{General} and \textit{Crowding}, which more frequently depict people, user contexts, and market-relevant scenes. For T.V.---where technical completeness is pivotal---both agents achieve generally higher scores in \textit{Technical Features} and \textit{Visual Condition}, where addressing salient technical flaws and visual constraints is a key challenge. Meanwhile, Innov. and C.A. achieve high scores not only in these technology-intensive domains but also in \textit{Spatial Layout}, which offers opportunities for creative mobile applications across diverse industries. These domain-aligned trends support the coherence and reliability of the benchmark design and evaluation framework.

\subsubsection{Modality Ablation}
Figure~\ref{fig:f2} analyzes the impact of different modality configurations on this task. We consider four model groups: a caption-only baseline, an open-source MLLM, GPT-5, and our agents, with the latter three receiving multimodal inputs. Across nearly all metrics, the caption baseline often lacks critical visual cues and therefore performs poorly in all domains, with the exception of \textit{General}, whose simpler, everyday scenes are more easily verbalized. Meanwhile, the multimodal baseline substantially improves overall performance over the caption baseline, especially in the image-specific domains. These findings demonstrate that text alone cannot fully capture visual details and is therefore insufficient for effective business ideation. With multimodal inputs, our agents achieve substantial gains in the two creativity-related metrics through GRPO, while SFT alone already improves the four feasibility-related metrics.

\begin{figure}[!t]
  \centering
  \includegraphics[width=\linewidth]{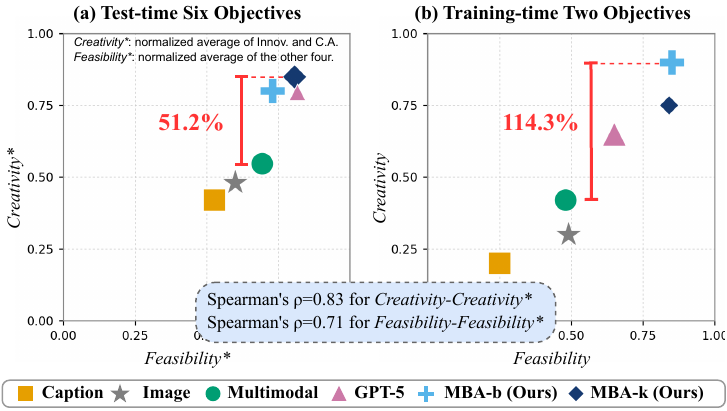}
  \caption{Plot of creativity--feasibility scores. \textbf{(a)} aggregates \textit{creativity$^*$} and \textit{feasibility$^*$} from the six test-time metrics, while \textbf{(b)} uses the two training-time rewards. Rankings are consistent across panels, with Spearman rank correlations of $\rho=0.83$ and $\rho=0.71$, respectively.}
  \label{fig:f5}
\end{figure}

\subsubsection{Validating Training-time Rewards}
Figure~\ref{fig:f5} examines how well our two training-time objectives align with the six evaluation criteria, using cross-model performance trends. We group the six metrics into two counterparts: Innov. and C.A. for \textit{Creativity*}, and the remaining four for \textit{Feasibility*}, mirroring our \textit{Creativity} and \textit{Feasibility}, respectively. We further extend the four model groups with an image-only baseline for broader comparison. As a result, model rankings show similar distributions across both metrics, with our agents leading the multimodal baseline by 51.2\% and 114.3\%, respectively. This trend is also confirmed by Spearman's $\rho$ of 0.83 and 0.71, computed across the model groups. This suggests that our two rewards generalize robustly across rubrics oriented toward both original and viable ideation.

\subsubsection{Cross-modal Mismatch}
\begin{figure}[!t]
  \centering
  \includegraphics[width=\linewidth]{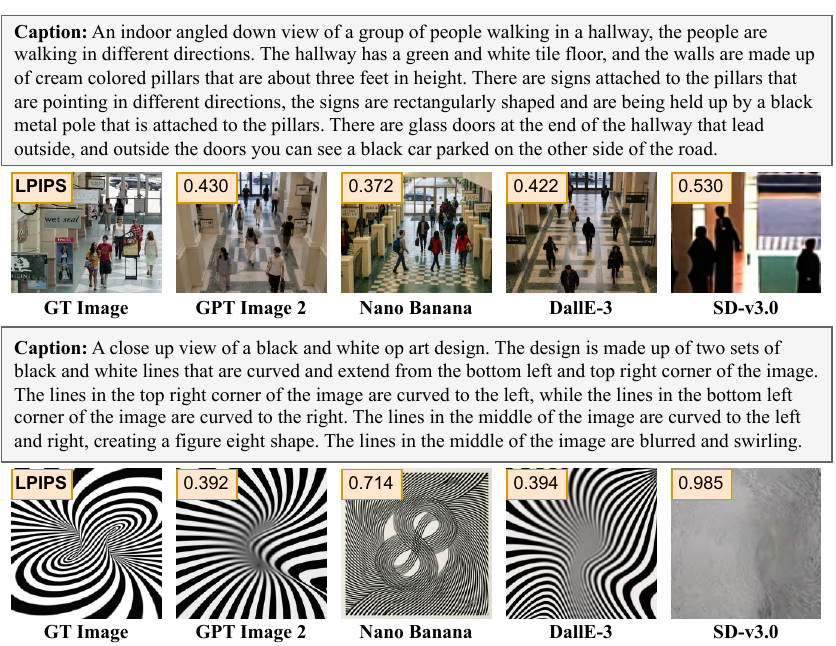}
  \caption{Qualitative examples of modality mismatch across four strong generative models, where visual details difficult to verbalize yield high LPIPS scores relative to the GT image (lower LPIPS indicates greater perceptual similarity).}
  \label{fig:f6}
\end{figure}

Figure~\ref{fig:f6} qualitatively analyzes modality mismatch using strong generative models~\cite{gpt4o,gemini,sd}. Each image shows its LPIPS distance from the original; higher is worse. The upper panel illustrates how detailed captions fail to preserve fine-grained visual details, yielding poor LPIPS scores (\eg, misplaced signage). The bottom panel shows that even accurate captions cannot reconstruct unverbalizable visual semantics (\eg, a swirling 3D pattern at the center), again producing poor scores. These failures reveal that each modality carries information the other cannot capture---a gap pervasive in real-world scenes. Such cross-modal translation can distort user intent, motivating multimodal business ideation.


\section{Limitations \& Future Work}
Beyond patent-constrained business ideation, MBA lays the groundwork for broader multimodal business ideation while leaving three fundamental limitations to be addressed for more effective real-world deployment. These challenges point to promising directions for future research.

First, MBA currently focuses on image and text, although real-world environments also contain informative audio, olfactory, tactile, and other sensory signals. Just as visual information can reveal business-relevant factors that are difficult to express in text, these modalities may expose additional latent needs and opportunities. For example, vocal tone and background sound may indicate emotion or regional characteristics, while olfactory cues may be critical in food, healthcare, manufacturing, and safety applications. Extending MBA to richer sensory inputs could therefore improve the diversity and contextual specificity of generated ideas.

Second, MBA reasons over spatial image--text inputs and does not explicitly model temporal information. In this respect, we assume that videos can further provide motion, behavioral, and causal context that is absent from a single frame, leading to substantially different business opportunities. For instance, a static image of roadside vehicles may ideate or suggest parking-related services, whereas a video showing a child abruptly emerging between them reveals a safety risk and thus motivates preventive solutions. Therefore, temporal multimodal reasoning is a promising research direction for generating more robust and actionable ideas.

Third, MBA generates and evaluates ideas independently of the prospective entrepreneur. In practice, feasibility depends on various factors such as available capital, expertise, location, social network, and risk tolerance. The same idea may be appropriate for one user but unrealistic for another. Collecting and analyzing such personalized information is still challenging, yet it is also essential for advancing toward real-world business ideation. Future work should therefore condition both generation and evaluation on user-specific profiles, enabling personalized business ideation that is not only creative and feasible in general, but realistically executable by a particular individual.

\section{Conclusion}
We introduced MBA-Bench, the first multimodal benchmark for business ideation, comprising 30K multimodal samples across six domains with up to eight metrics. We further proposed two task-specialized agents for the blind and known evaluation settings through SFT--GRPO training. Both consistently outperform diverse baselines while remaining competitive with strong closed-source MLLMs. We hope our work inspires further academic research on multimodal business ideation and broadens access to opportunity discovery beyond text across physical environments, digital interfaces, and everyday settings for aspiring entrepreneurs at all levels.

\bibliography{aaai2027}


\clearpage

\let\addcontentsline\savedaddcontentsline


\definecolor{cvprblue}{rgb}{0.21,0.49,0.74}

\setcounter{section}{0}
\setcounter{subsection}{0}
\setcounter{figure}{0}
\setcounter{table}{0}
\setcounter{equation}{0}

\renewcommand{\thesection}{\Alph{section}}
\renewcommand{\thetable}{A\arabic{table}}
\renewcommand{\thefigure}{A\arabic{figure}}
\renewcommand{\theequation}{A\arabic{equation}}

\startcontents[supplement]


\begingroup
\color{cvprblue}
\printcontents[supplement]{}{1}{%
    \section*{\color{black}Contents}
    \setcounter{tocdepth}{2}
}
\endgroup


\newpage

\section{Experimental Details}

\subsection{Device Information}
All experiments were conducted on Ubuntu 22.04 LTS using eight NVIDIA RTX A6000 GPUs and dual AMD EPYC 7513 CPUs. Training used approximately 45 GiB of GPU memory per device, while evaluation used approximately 30 GiB. We used Python 3.10.13, PyTorch 2.2.0 with CUDA 12.1, and a random seed of 2026.

\subsection{Implementation Details}
We use PaliGemma2~\cite{captioner} as the long-context image captioner and GPT-4o~\cite{gpt4o} to construct MBA-Bench. Our MBA agents are initialized from Qwen2.5-VL-7B-Instruct~\cite{qwen} and trained through LoRA-based supervised fine-tuning followed by GRPO~\cite{deepseek}. Qwen2.5-VL-72B-Instruct is employed as the reward judge during training, whereas InternVL2.5-78B~\cite{internvl} is used for evaluation. The feasibility reward combines market relevance, measured using FAISS~\cite{faiss}, and factuality, assessed using FActScore~\cite{factscore}. We set the random seed to 2026 for reproducibility, and all reported results are obtained from a single run.

\subsection{Evaluation Details}
We evaluate all models on the MBA-Bench test set using the same multimodal inputs and unified question prompt. Following the PBIG evaluation protocol~\cite{pbig}, each generated business idea is assessed by InternVL2.5-78B~\cite{internvl} as an MLLM judge across six business-oriented dimensions, with score ranges of 1--4, 1--4, 1--5, 1--4, 0--3, and 0--3, respectively. These metrics are selected to jointly capture the feasibility of an idea---including its clarity, technical soundness, user demand, and market potential---and its creativity in terms of novelty and differentiation. We report the mean and standard deviation of each metric across test images. Detailed evaluation reward statistics are provided in Table~\ref{tab:evaluation_rewards}.

In this work, we adopt the six PBIG metrics~\cite{pbig}, which jointly assess whether a business idea is concrete, technically feasible, innovative, competitively differentiated, demand-driven, and commercially scalable. This balanced design avoids rewarding novelty alone and instead captures originality, feasibility, and market relevance. We evaluate these dimensions using an MLLM-as-a-judge protocol, as image-grounded business ideas require joint reasoning over visual evidence, technical plausibility, and market context beyond lexical similarity. For reliability and comparability, all models are evaluated with the same fixed judge and rubric, each dimension is scored independently, and the evaluation judge is separated from the training reward model.

\subsection{Hyperparameters}
For LoRA-based adaptation~\cite{lora}, we use a rank of 32, a scaling factor of 64, and a dropout rate of 0.05. SFT is performed for two epochs with a learning rate of $2\times10^{-5}$, a warmup ratio of 0.1, and a global batch size of 32, followed by one epoch of GRPO~\cite{deepseek} with a learning rate of $1\times10^{-6}$, four sampled responses per prompt, a KL coefficient of 0.02, and a batch size of 4. For MBA-k, the reward weights for \textit{Specificity}, \textit{Technical Validity}, \textit{Innovativeness}, \textit{Competitive Advantage}, \textit{Need Validity}, \textit{Market Size}, \textit{Creativity}, and \textit{Feasibility} are set to 0.12, 0.12, 0.20, 0.16, 0.12, 0.08, 0.10, and 0.10, respectively, whereas MBA-b uses creativity and feasibility weights of 0.70 and 0.30.

\subsection{Pseudocode}

\begin{algorithm}[tb]
\caption{Construction of MBA-Bench}
\label{alg:mba_bench}
\textbf{Input}: Domain datasets $\{\mathcal{D}_d\}_{d\in\mathcal{D}}$,
captioner $C$, generator $G$, web retriever $R$, and question set $\mathcal{Q}$\\
\textbf{Output}: MBA-Bench training and test sets
$\mathcal{B}_{\mathrm{tr}}$ and $\mathcal{B}_{\mathrm{te}}$
\begin{algorithmic}[1]
\STATE Initialize $\mathcal{B}\gets\emptyset$
\FOR{each domain $d\in\mathcal{D}$}
    \STATE Select representative images $\mathcal{I}_d$ using domain-specific criteria
    \FOR{each image $v\in\mathcal{I}_d$}
        \STATE Generate caption $c\gets C(v)$
        \STATE Extract retrieval query $q\gets G_{\mathrm{query}}(v,c,d)$
        \STATE Retrieve market evidence $e\gets R(q)$
        \FOR{each business question $b\in\mathcal{Q}$}
            \STATE Construct unified prompt $x\gets(v,c,d,b,q,e)$
            \STATE Generate $K$ reference ideas $\mathcal{Y}_x\gets G_{\mathrm{idea}}(x,K)$
            \STATE Add $\{(x,y)\mid y\in\mathcal{Y}_x\}$ to $\mathcal{B}$
        \ENDFOR
    \ENDFOR
\ENDFOR
\STATE Split $\mathcal{B}$ by image into
$\mathcal{B}_{\mathrm{tr}}$ and $\mathcal{B}_{\mathrm{te}}$ at a $95{:}5$ ratio
\STATE \textbf{return} $\mathcal{B}_{\mathrm{tr}},\mathcal{B}_{\mathrm{te}}$
\end{algorithmic}
\end{algorithm}

\begin{algorithm}[tb]
\caption{SFT--GRPO Training of an MBA Agent}
\label{alg:mba_training}
\textbf{Input}: Training set $\mathcal{B}_{\mathrm{tr}}$, base MLLM $M_0$,
reward judge $J$, MBA-Library $\mathcal{L}$, setting
$s\in\{\mathrm{b},\mathrm{k}\}$, reward set $\mathcal{M}_s$,
and weights $\{\sigma_m^{(s)}\}$\\
\textbf{Output}: Trained policy $\pi_s$
\begin{algorithmic}[1]
\STATE $\pi_{\theta,\phi}\gets\mathrm{LoRA}(M_0)$
\STATE Train $\pi_{\theta,\phi}$ on question--reference pairs using
$\mathcal{L}_{\mathrm{SFT}}
=-\sum_t\log p_{\theta,\phi}(y_t\mid x,y_{<t})$
\STATE $\pi_s\gets\pi_{\mathrm{SFT}}$,\quad
$\pi_{\mathrm{ref}}\gets\mathrm{stopgrad}(\pi_{\mathrm{SFT}})$
\FOR{each training prompt $x$}
    \STATE $\{o_i\}_{i=1}^{G}\sim\pi_s(\cdot\mid x)$
    \STATE $(R_{c,1:G},R_{f,1:G})
    \gets\mathrm{Reward}(o_{1:G};J,\mathcal{L})$
    \IF{$s=\mathrm{k}$}
        \STATE $\{R_{m,1:G}\}_{m\in\mathcal{E}}
        \gets\mathrm{EvalReward}(o_{1:G};J,\mathcal{E})$
    \ENDIF
    \STATE $r_i^{(s)}\gets
    \sum_{m\in\mathcal{M}_s}\sigma_m^{(s)}R_{m,i}$
    \STATE $A_i\gets\mathrm{GroupNormalize}(r_{1:G}^{(s)})$
    \hfill $\triangleright$ Eq.~(\ref{eq:grpo_advantage})
    \STATE Update $\pi_s$ by minimizing $\mathcal{L}_{\mathrm{GRPO}}$
    \hfill $\triangleright$ Eq.~(\ref{eq:grpo_objective})
\ENDFOR
\STATE \textbf{return} $\pi_s$
\end{algorithmic}
\end{algorithm}

\begin{algorithm}[tb]
\caption{Evaluation on MBA-Bench}
\label{alg:mba_evaluation}
\textbf{Input}: $\{\mathcal{B}_v\}_{v\in\mathcal{V}_{\mathrm{te}}}$,
models $\mathcal{F}$, judge $J_{\mathrm{eval}}$, and metrics
$\mathcal{E}$\\
\textbf{Output}: $\{(\boldsymbol{\mu}_f,
\boldsymbol{\sigma}_f)\}_{f\in\mathcal{F}}$
\begin{algorithmic}[1]
\FOR{$f\in\mathcal{F}$}
    \STATE $\mathcal{S}_f\gets\emptyset$
    \FOR{$v\in\mathcal{V}_{\mathrm{te}}$}
        \STATE $\hat{\mathcal{Y}}_{f,v}\gets
        \{f(x)\mid x\in\mathcal{B}_v\}$
        \STATE $\mathcal{R}_{f,v}\gets
        \{J_{\mathrm{eval}}(x,\hat{y};\mathcal{E})
        \mid (x,\hat{y})\in
        (\mathcal{B}_v,\hat{\mathcal{Y}}_{f,v})\}$
        \STATE $\bar{\mathbf{s}}_{f,v}\gets
        \operatorname{Mean}(\mathcal{R}_{f,v})$
        \STATE $\mathcal{S}_f\gets
        \mathcal{S}_f\cup\{\bar{\mathbf{s}}_{f,v}\}$
    \ENDFOR
    \STATE $(\boldsymbol{\mu}_f,\boldsymbol{\sigma}_f)
    \gets\operatorname{MeanStd}(\mathcal{S}_f)$
\ENDFOR
\STATE \textbf{return}
$\{(\boldsymbol{\mu}_f,\boldsymbol{\sigma}_f)\}_{f\in\mathcal{F}}$
\end{algorithmic}
\end{algorithm}

\subsubsection{MBA-Bench Construction}
Algorithm~\ref{alg:mba_bench} summarizes the construction pipeline of MBA-Bench. For each domain $d$, we select the top $N_d$ images according to an annotation-derived relevance score $s_d$ and generate a caption for each selected image. We then extract a visually grounded retrieval query and collect the corresponding market evidence through web retrieval. For each of the three business questions, the image, caption, domain, query, and evidence are combined into a unified prompt from which $K=5$ reference ideas are generated. This procedure yields $(\sum_d N_d)|\mathcal{Q}|K=2{,}000\times3\times5=30{,}000$ samples, which are split by image into training and test sets at a ratio of $95{:}5$ to prevent data overlap.


\begin{table*}[t]
\centering
\footnotesize
\setlength{\tabcolsep}{6pt}
\renewcommand{\arraystretch}{1.10}

\begin{tabular}{
    @{}
    >{\raggedright\arraybackslash}p{0.13\textwidth}
    >{\raggedright\arraybackslash}p{0.39\textwidth}
    >{\raggedright\arraybackslash}p{0.415\textwidth}
    @{}
}
\toprule
\textbf{Component}
&
\textbf{Description}
&
\textbf{Example}
\\
\midrule

\rowcolor{black!5}
\textbf{Domain}
&
The visual domain assigned to an image, together with a
domain-specific focus that identifies the visual attributes and
business opportunities relevant to ideation.
&
\textbf{Aesthetic / Texture / Material Semantics}: Identify startup
opportunities from texture, material quality, surface appearance,
tactile or visual differentiation, design, and product experience.
\\
\addlinespace[3pt]

\textbf{Image}
&
The original visual input selected from a public source dataset using
domain-specific annotation criteria. It serves as the primary source
of visual evidence throughout business ideation.
&
A close-up DTD image of a painted canvas containing smeared blue,
white, red, and black pigments, including black streaks and a small
red triangular mark near the bottom-right corner.
\\
\addlinespace[3pt]

\rowcolor{black!5}
\textbf{Caption}
&
An automatically generated natural-language description that provides
auxiliary semantic context for the image. The caption may summarize
visible objects, materials, colors, spatial relations, and surface
properties, but does not replace the original image.
&
``A close-up view of a painting on a canvas. The painting is made up
of different shades of blue, white, and red paint. Blue and white
paint is smeared across the canvas, together with smeared black lines
and a small red triangle in the bottom-right corner.''
\\
\addlinespace[3pt]

\textbf{Query}
&
A visually grounded web-search query generated for a specific
image--business-question pair. It translates the visual context and
business objective into a market-oriented information need.
&
``Sustainable materials for interior design cost reduction.''
\\
\addlinespace[3pt]

\rowcolor{black!5}
\textbf{Evidence}
&
Market-grounding information retrieved using the generated query.
It may describe relevant customer needs, industry practices,
technologies, sustainability opportunities, economic considerations,
or quantitative market facts.
&
``Eco-friendly interiors can incorporate LED lighting,
energy-efficient appliances, low-VOC paints, and recycled materials.
Early coordination between construction and interior design can
reduce costly design changes and avoid unnecessary rework, while
durable materials and efficient layouts can improve long-term
functionality and property value.''
\\
\addlinespace[3pt]

\textbf{Business Question}
&
One of three business-oriented lenses used to guide ideation toward
cost and resource efficiency, intelligent technology and automation,
or customer experience and business growth.
&
\textbf{Cost and Resource Efficiency}: Focus on cost reduction,
resource productivity, waste reduction, energy or material efficiency,
and sustainability-linked savings.
\\

\bottomrule
\end{tabular}

\caption{Components of the unified MBA-Bench prompt and an example derived from the \textit{Shape \& Texture} domain.}
\label{tab:mba_bench}
\end{table*}

\subsubsection{MBA-Agent Training}
Algorithm~\ref{alg:mba_training} presents the two-stage optimization procedure for MBA agents. We first perform LoRA-based supervised fine-tuning on the question--reference idea pairs to obtain initial policy $\pi_{\mathrm{SFT}}$. From this checkpoint, GRPO samples $G$ candidate ideas for each unified prompt and assigns group-relative rewards. Creativity is measured relative to the reference ideas, whereas feasibility is computed from market relevance and factuality using MBA-Library. MBA-b optimizes only creativity and feasibility, while MBA-k additionally incorporates the six disclosed evaluation criteria. The weighted rewards are normalized within each group and used to update the policy under KL regularization toward the frozen SFT reference.

\subsubsection{MBA-Bench Evaluation}
Algorithm~\ref{alg:mba_evaluation} summarizes the evaluation procedure on MBA-Bench. Each model generates one business idea for every unified test prompt, and the evaluation judge assigns a six-dimensional score vector covering \textit{Specificity}, \textit{Technical Validity}, \textit{Innovativeness}, \textit{Competitive Advantage}, \textit{Need Validity}, and \textit{Market Size}. The 15 instance-level scores associated with each image are first averaged, after which the mean and standard deviation are computed across the 100 image-level results.


\section{Dataset Details}

\subsection{MBA-Bench}
MBA-Bench organizes each sample around six complementary components, as summarized in Tables~\ref{tab:mba_bench} and~\ref{tab:dataset_statistics}. The image provides the primary visual evidence, while the domain specifies the visual and business context under which that evidence should be interpreted. The caption supplies auxiliary semantic information about salient objects, materials, attributes, and spatial relations. The business question defines the ideation objective, such as improving cost efficiency, enabling technology-driven solutions, or enhancing user experience. Based on the image and question, the retrieval query converts the visual context into a market-oriented information need, and the retrieved evidence grounds generation in relevant industry practices, customer needs, and market facts. Together, these components form the unified question prompt used to generate structured reference ideas. The following subsections introduce the six source datasets and explain how each supports its corresponding visual domain.

\subsubsection{ADE20K}
ADE20K~\cite{ade20k} supplies 500 images for the \textit{General} domain. Its diverse indoor and outdoor scenes contain broad combinations of everyday objects and environments, providing general-purpose visual contexts that can be comparatively well represented through captions. We rank images by the diversity of their annotated object categories and retain semantically rich scenes. This domain serves as a broadly verbalizable reference point against which the more visually implicit domains can be compared.

\begin{table*}[t]
\centering
\footnotesize
\setlength{\tabcolsep}{2.3pt}
\renewcommand{\arraystretch}{1.10}

\begin{adjustbox}{width=\textwidth,center}
\begin{tabular}{@{}llcccccccc@{}}
\toprule
\textbf{Comparator}
& Metric
& MBA-k
& Competitor Mean
& $\Delta$
& $n_{\mathrm{eff}}$
& Wilcoxon ($W$)
& Raw $p$
& Holm $p$
& Sig. Diff.
\\
\midrule

\rowcolor{black!5}
\textbf{GPT-5 mini}
& \textit{Specificity}
& 3.980 & 4.000 & -0.020 & 25
& 0.0
& $3.06\!\times\!10^{-6}$
& $3.06\!\times\!10^{-6}$
& Yes \\

\rowcolor{black!5}
& \textit{Technical Validity}
& 2.998 & 3.065 & -0.067 & 51
& 0.0
& $2.92\!\times\!10^{-10}$
& $5.83\!\times\!10^{-10}$
& Yes \\

\rowcolor{black!5}
& \textit{Innovativeness}
& 3.999 & 3.586 & 0.413 & 97
& 0.0
& $1.14\!\times\!10^{-17}$
& $4.55\!\times\!10^{-17}$
& Yes \\

\rowcolor{black!5}
& \textit{Competitive Advantage}
& 3.321 & 3.004 & 0.317 & 98
& 0.0
& $7.40\!\times\!10^{-18}$
& $3.70\!\times\!10^{-17}$
& Yes \\

\rowcolor{black!5}
& \textit{Need Validity}
& 2.937 & 2.768 & 0.169 & 81
& 128.0
& $4.56\!\times\!10^{-13}$
& $1.37\!\times\!10^{-12}$
& Yes \\

\rowcolor{black!5}
& \textit{Market Size}
& 2.751 & 2.075 & 0.676 & 100
& 0.0
& $3.65\!\times\!10^{-18}$
& $2.19\!\times\!10^{-17}$
& Yes \\
\addlinespace[3pt]

\textbf{Gemini 3.1 Pro}
& \textit{Specificity}
& 3.980 & 4.000 & -0.020 & 25
& 0.0
& $3.06\!\times\!10^{-6}$
& $9.18\!\times\!10^{-6}$
& Yes \\

& \textit{Technical Validity}
& 2.998 & 3.040 & -0.042 & 40
& 12.5
& $4.60\!\times\!10^{-8}$
& $1.84\!\times\!10^{-7}$
& Yes \\

& \textit{Innovativeness}
& 3.999 & 3.985 & 0.015 & 26
& 42.0
& $2.90\!\times\!10^{-4}$
& $5.80\!\times\!10^{-4}$
& Yes \\

& \textit{Competitive Advantage}
& 3.321 & 3.006 & 0.315 & 98
& 0.0
& $7.43\!\times\!10^{-18}$
& $3.72\!\times\!10^{-17}$
& Yes \\

& \textit{Need Validity}
& 2.937 & 2.969 & -0.032 & 52
& 346.5
& $1.45\!\times\!10^{-3}$
& $1.45\!\times\!10^{-3}$
& Yes \\

& \textit{Market Size}
& 2.751 & 2.347 & 0.403 & 100
& 2.5
& $3.92\!\times\!10^{-18}$
& $2.35\!\times\!10^{-17}$
& Yes \\
\addlinespace[3pt]

\rowcolor{black!5}
\textbf{InternVL2.5-8B}
& \textit{Specificity}
& 3.980 & 3.625 & 0.355 & 98
& 0.0
& $7.66\!\times\!10^{-18}$
& $2.12\!\times\!10^{-17}$
& Yes \\

\rowcolor{black!5}
& \textit{Technical Validity}
& 2.998 & 3.008 & -0.010 & 26
& 108.5
& $7.82\!\times\!10^{-2}$
& $7.82\!\times\!10^{-2}$
& No \\

\rowcolor{black!5}
& \textit{Innovativeness}
& 3.999 & 3.195 & 0.804 & 100
& 0.0
& $3.15\!\times\!10^{-18}$
& $1.89\!\times\!10^{-17}$
& Yes \\

\rowcolor{black!5}
& \textit{Competitive Advantage}
& 3.321 & 2.948 & 0.373 & 99
& 0.0
& $5.31\!\times\!10^{-18}$
& $2.12\!\times\!10^{-17}$
& Yes \\

\rowcolor{black!5}
& \textit{Need Validity}
& 2.937 & 2.389 & 0.547 & 100
& 0.0
& $3.67\!\times\!10^{-18}$
& $1.89\!\times\!10^{-17}$
& Yes \\

\rowcolor{black!5}
& \textit{Market Size}
& 2.751 & 2.125 & 0.626 & 99
& 0.0
& $5.45\!\times\!10^{-18}$
& $2.12\!\times\!10^{-17}$
& Yes \\

\bottomrule
\end{tabular}
\end{adjustbox}

\caption{
Paired two-sided Wilcoxon signed-rank tests comparing MBA-k, with GPT-5 mini, Gemini 3.1 Pro, and
InternVL2.5-8B on the same 100-image test set. For each model and metric,
the 15 responses per image were averaged, yielding 100 paired
image-level observations. MBA-k and \emph{Competitor Mean}
report the means of our model and the comparator named in each block,
respectively. $\Delta=\bar{x}_{\text{MBA-k}}
-\bar{x}_{\mathrm{competitor}}$, where positive values favor
MBA-k. $n_{\mathrm{eff}}$ is the number of non-zero paired
differences, with zero differences excluded following the Wilcoxon
convention. Holm correction was applied across the six metrics within
each comparator, and adjusted $p<0.05$ indicates statistical
significance.
}
\label{tab:wilcoxon}
\end{table*}

\subsubsection{RICO}
RICO~\cite{rico} contributes 350 mobile-interface screenshots to the \textit{Spatial Layout} domain. We prioritize screenshots with large numbers of interface components and substantial textual content, yielding visually dense layouts with varied controls, menus, and information structures. These samples are included because spatial organization and component relationships can reveal opportunities related to interface usability, service design, and user engagement that are difficult to infer from isolated textual descriptions.

\subsubsection{MS-COCO}
MS-COCO~\cite{coco} provides 350 images for the \textit{Crowding} domain. We restrict the candidate pool to images containing at least ten annotated people and rank them using person count and object diversity. The resulting scenes represent dense human activity in public, commercial, and social environments, supporting business ideation involving customer flow, capacity management, safety, accessibility, and resource allocation.

\subsubsection{VisA}
VisA~\cite{visa} contributes 350 industrial images to the \textit{Visual Condition} domain, using a fixed sampling ratio that favors anomalous examples while retaining normal cases for comparison. Its fine-grained surface irregularities and manufacturing defects provide visual signals that may be difficult to express completely in captions. We include VisA to support opportunities related to quality inspection, predictive maintenance, and defect-aware automation.

\subsubsection{DTD}
DTD~\cite{dtd} supplies 350 close-up texture images for the \textit{Shape \& Texture} domain. We sample images to maintain approximately uniform coverage across texture attributes, avoiding overrepresentation of visually frequent categories. These images emphasize material appearance, surface patterns, color composition, and tactile impressions, enabling ideation for product design, fashion, interiors, digital assets, and material-oriented applications.

\subsubsection{DeepPCB}
DeepPCB~\cite{deeppcb} provides 100 printed-circuit-board images for the \textit{Technical Features} domain, with priority given to samples containing annotated defects. The images contain small components, repeated structures, conductive traces, and localized manufacturing irregularities that require fine-grained visual interpretation. This dataset supports technical business opportunities involving automated inspection, manufacturing diagnostics, repair assistance, reliability monitoring, and electronics production.

\subsection{MBA-Library}
MBA-Library is an external knowledge base used to ground the feasibility reward in retrieved evidence rather than relying solely on an MLLM judge. It integrates scientific literature, structured entities, and Wikipedia-based evidence to provide complementary technical, commercial, and factual knowledge. OpenAlex and Wikidata constitute its primary knowledge sources, while FAISS and FActScore support market-relevance retrieval and factuality verification, respectively. All constituent resources and implementations are used in accordance with their respective licenses and terms of use.

\subsubsection{OpenAlex}
OpenAlex~\cite{openalex} provides an open index of scholarly works, authors, venues, institutions, and research concepts. We use its records to connect generated ideas with existing technologies, scientific developments, and implementation evidence, thereby strengthening the technical grounding of feasibility assessment. This dataset is released under the CC0 public-domain dedication.

\subsubsection{Wikidata}
Wikidata~\cite{wikidata} contributes structured entities and relations covering technologies, organizations, products, industries, and locations. This entity-centric knowledge complements unstructured documents by explicitly representing relationships among real-world concepts, improving coverage of relevant technologies and commercial ecosystems. Wikidata's structured data is released under the CC0 public-domain dedication.

\subsubsection{FAISS Library}
FAISS~\cite{faiss} is used as the vector-search library for indexing dense representations of MBA-Library records and retrieving the top-$k$ entries most similar to each generated idea. Market relevance is computed from the similarity between the idea and the retrieved evidence and normalized to $[0,1]$ as one component of the feasibility reward. The official FAISS implementation is distributed under the MIT License.

\subsubsection{FActScore}
FActScore~\cite{factscore} is adapted to measure whether factual claims in a generated idea are supported by external knowledge. The idea is decomposed into atomic claims, which are verified against retrieved Wikipedia passages, and their aggregated support is normalized to $[0,1]$ as the factuality component of the feasibility reward. This implementation is distributed under the MIT License.


\section{Additional Quantitative Evaluation}

\subsection{Further Statistical Analysis}
As shown in Table~\ref{tab:wilcoxon}, MBA-k achieves statistically significant improvements in 12 of 18 model--metric comparisons. It significantly outperforms all three comparators in Innovativeness, Competitive Advantage, and Market Size, while also improving Need Validity over GPT-5 mini and InternVL2.5-8B and Specificity over InternVL2.5-8B. Conversely, GPT-5 mini and Gemini 3.1 Pro perform better on Specificity and Technical Validity, and Gemini 3.1 Pro also on Need Validity; the Technical Validity difference from InternVL2.5-8B is not significant. Overall, the alternative reward configuration strengthens novelty, differentiation, and market potential, with trade-offs in clarity and technical validity.


\section{Additional Ablation Studies}

\begin{table}[!t]
\centering

\begin{adjustbox}{max width=\columnwidth,center}
\begin{tabular}{@{}lcccccc@{}}
    \toprule
    Captioner
    & Spec.
    & T.V.
    & Innov.
    & C.A.
    & N.V.
    & M.S. \\
    \midrule

    Gemini
    & 2.66
    & 2.13
    & 1.95
    & \textbf{1.69}
    & \textbf{1.50}
    & 1.09 \\

    GPT-5
    & \textbf{2.69}
    & \textbf{2.15}
    & 1.97
    & 1.68
    & 1.49
    & \textbf{1.10} \\

    PaliGemma2-10B
    & 2.60
    & 2.08
    & \textbf{1.98}
    & 1.67
    & 1.48
    & 1.08 \\
    
    \bottomrule
\end{tabular}
\end{adjustbox}

\caption{
Caption-based performance using different image captioners.
Spec.: \textit{Specificity};
T.V.: \textit{Technical Validity};
Innov.: \textit{Innovativeness};
C.A.: \textit{Competitive Advantage};
N.V.: \textit{Need Validity};
and M.S.: \textit{Market Size}.
}
\label{tab:captioner_ablation}
\end{table}

\subsection{Additional Captioning Models}
We examine whether the choice of image captioner materially affects downstream business-ideation performance by replacing PaliGemma2~\cite{captioner} with Gemini and GPT-5 while keeping the ideation model and all remaining inputs fixed. As shown in Table~\ref{tab:captioner_ablation}, the three captioners achieve broadly comparable results across the six evaluation dimensions. Gemini and GPT-5 provide modest gains in \textit{Specificity} and \textit{Technical Validity}, whereas PaliGemma2 remains competitive in \textit{Innovativeness} and \textit{Competitive Advantage}, and the differences in \textit{Need Validity} and \textit{Market Size} are negligible. This limited and metric-dependent variation indicates that the caption-based results are not primarily attributable to a weak captioning model, but instead reflect the information bottleneck introduced when visual observations are converted into text. Given its competitive downstream performance, fixed public weights, deterministic local inference, and independence from proprietary API updates, we therefore adopt PaliGemma2 as a reproducible and practical captioner for constructing MBA-Bench.

\begin{table}[!t]
\centering
\small

\begin{adjustbox}{max width=\columnwidth,center}
\begin{tabular}{@{}lccc@{}}
\toprule
Model
& \multicolumn{2}{c}{\textit{Severe Failure (\%, $\downarrow$)}}
& \textit{Invalid Format (\%, $\downarrow$)} \\
\cmidrule(lr){2-3}
& \textit{Semantic}
& \textit{Technical}
& \\
\midrule

\multicolumn{4}{@{}l}{\textit{Closed-source models}} \\

\quad GPT-5 mini
& 0.00
& 0.00
& 0.07 \\

\quad Gemini 3.1 Pro
& 0.00
& 0.00
& 0.00 \\

\midrule
\multicolumn{4}{@{}l}{\textit{Open-source models}} \\

\quad Qwen2.5-VL-32B
& 5.13
& 0.67
& 99.80 \\

\quad LLaVA-OneVision-7B
& 12.60
& 3.93
& 61.33 \\

\quad InternVL2.5-8B
& 6.73
& 1.00
& 16.93 \\

\rowcolor{green!10}
\quad \textbf{MBA-k}
& \textbf{0.40}
& \textbf{0.40}
& \textbf{3.40} \\

\bottomrule
\end{tabular}
\end{adjustbox}

\caption{Severe-failure and invalid-format rates over 1,500 responses from 100 test images. Lower is better. \textit{Semantic} follows the rubric-defined low-score thresholds, while \textit{Technical} denotes the subset with \textit{Technical Validity} $\leq 2$.}
\label{tab:supp:failure_audit}
\end{table}

\subsection{Reliability and Failure Analysis}
Table~\ref{tab:supp:failure_audit} compares severe failure and invalid-format rates across closed- and open-source models. A severe semantic failure is defined as a score of at most 2 for \textit{Specificity}, \textit{Technical Validity}, \textit{Innovativeness}, or \textit{Competitive Advantage}, or at most 1 for \textit{Need Validity} or \textit{Market Size}; the \textit{Technical} column reports the subset attributable to \textit{Technical Validity}. MBA-k achieves the lowest semantic failure rate among the open-source models at 0.40\%, compared with 5.13--12.60\% for the other open-source baselines, while remaining close to the zero-failure rates of the closed-source models. It also reduces invalid-format outputs to 3.40\%, compared with 16.93--99.80\% for the other open-source models. These violations do not necessarily indicate an inability to generate meaningful ideas; they primarily reflect incomplete or non-conforming four-field JSON outputs. The improved format reliability of MBA-k is consistent with the benefit of task-specific supervised fine-tuning on structured question--idea pairs, although we do not attribute the improvement solely to SFT without a dedicated ablation. All of its severe semantic failures arise from \textit{Technical Validity}, with no severe failures observed in the other five PBIG dimensions. Overall, MBA-k substantially closes the semantic and structured-output reliability gap between open- and closed-source models.


\section{Additional Qualitative Results}
\begin{figure*}[!p]
  \centering
  \includegraphics[
    width=\textwidth,
    height=\textheight,
    keepaspectratio
  ]{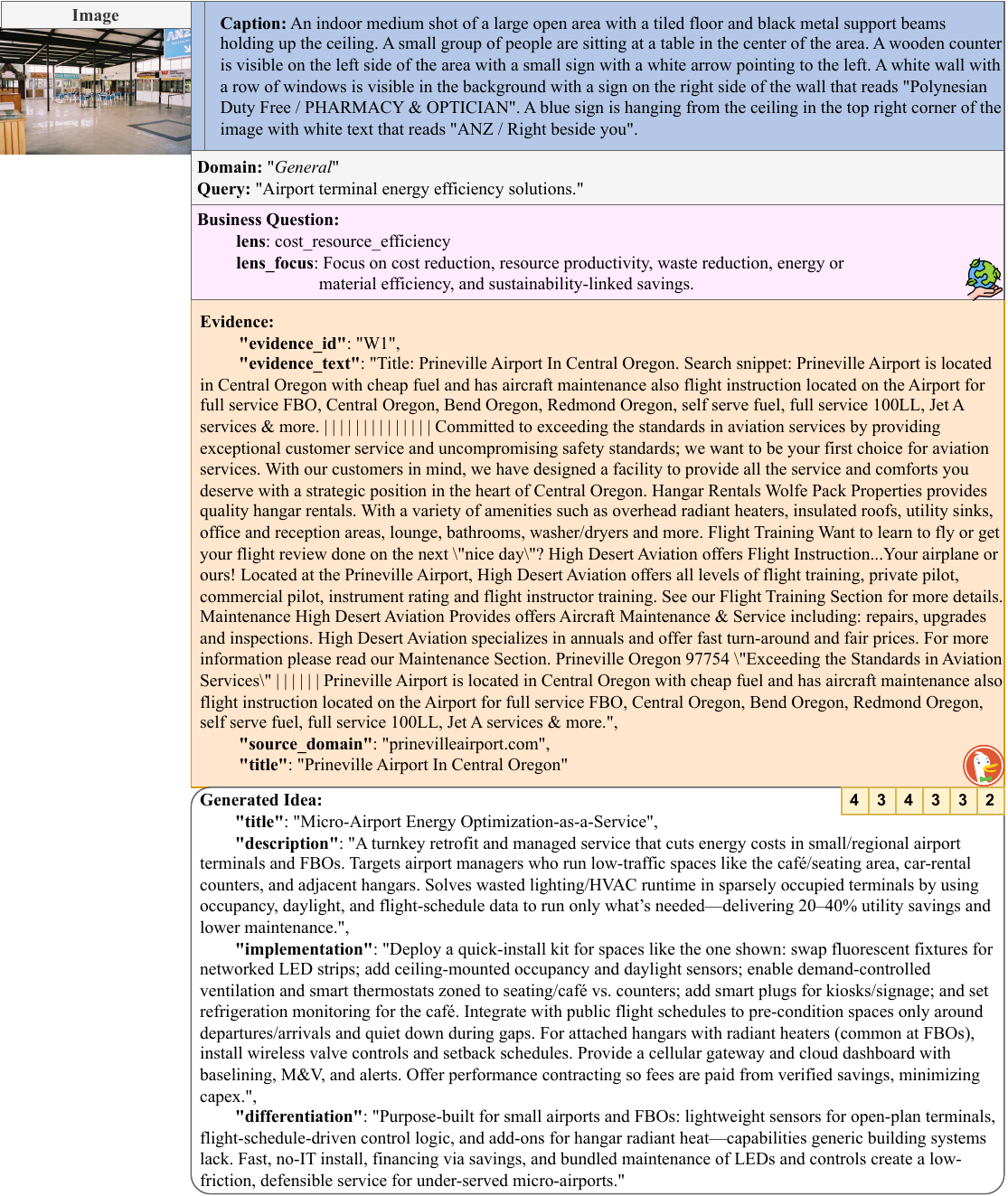}
  \caption{An example from the \textit{General} domain in MBA-Bench.}
  \label{fig:mba_general}
\end{figure*}

\begin{figure*}[!p]
  \centering
  \includegraphics[
    width=\textwidth,
    height=\textheight,
    keepaspectratio
  ]{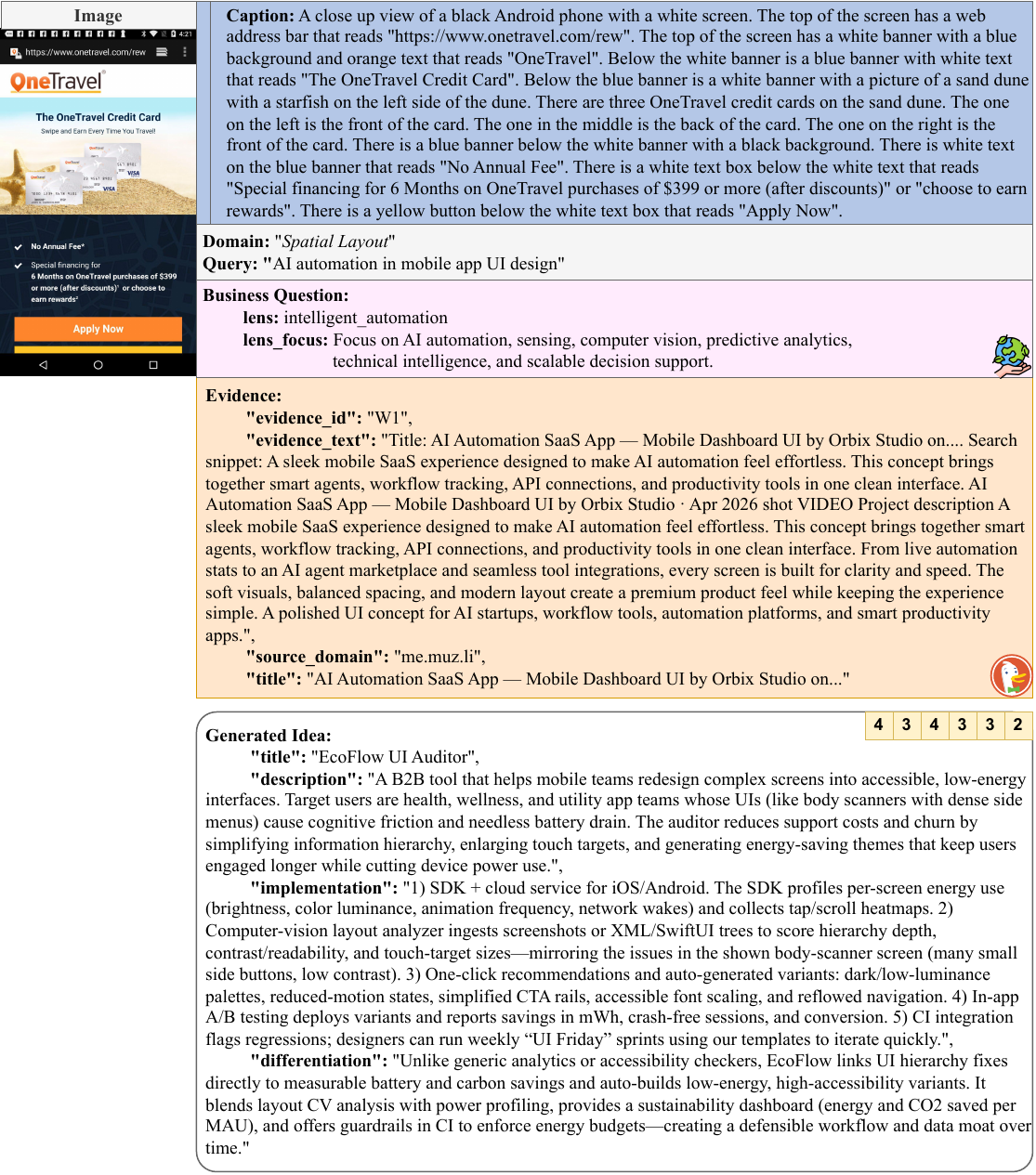}
  \caption{An example from the \textit{Spatial Layout} domain in MBA-Bench.}
  \label{fig:mba_spatial}
\end{figure*}

\begin{figure*}[!p]
  \centering
  \includegraphics[
    width=\textwidth,
    height=\textheight,
    keepaspectratio
  ]{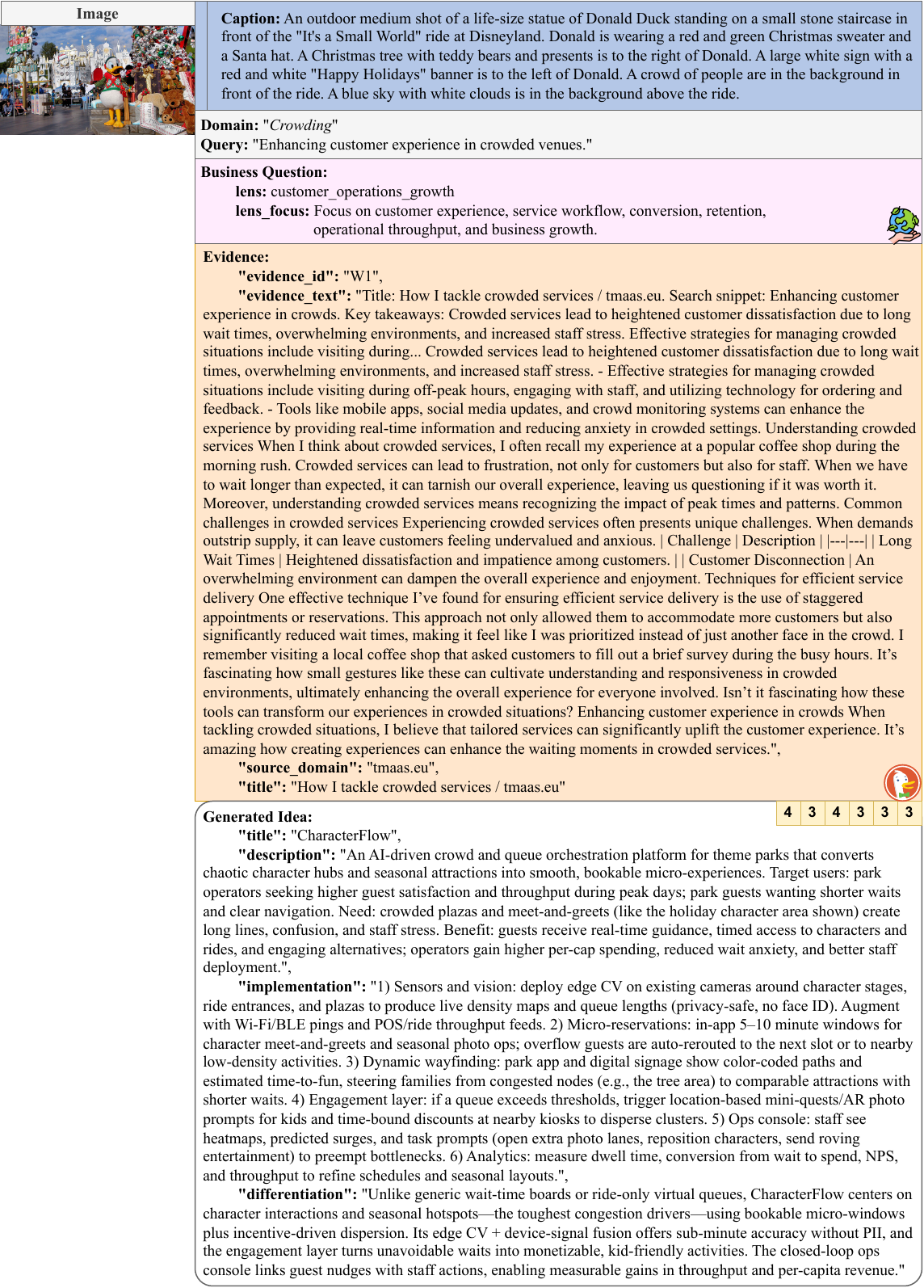}
  \caption{An example from the \textit{Crowding} domain in MBA-Bench.}
  \label{fig:mba_crowd}
\end{figure*}

\begin{figure*}[!p]
  \centering
  \includegraphics[
    width=\textwidth,
    height=\textheight,
    keepaspectratio
  ]{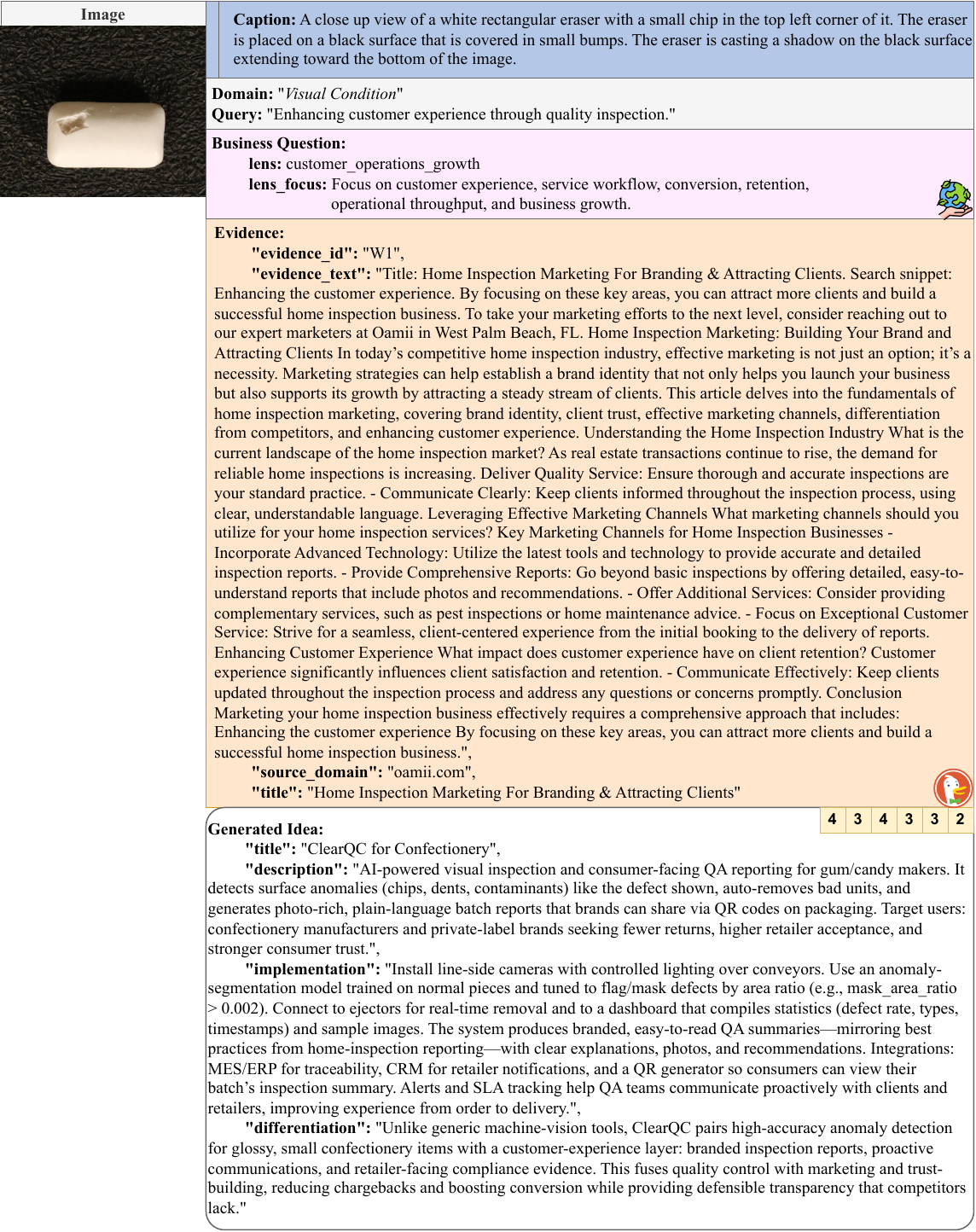}
  \caption{An example from the \textit{Visual Condition} domain in MBA-Bench.}
  \label{fig:mba_visual}
\end{figure*}

\begin{figure*}[!p]
  \centering
  \includegraphics[
    width=\textwidth,
    height=\textheight,
    keepaspectratio
  ]{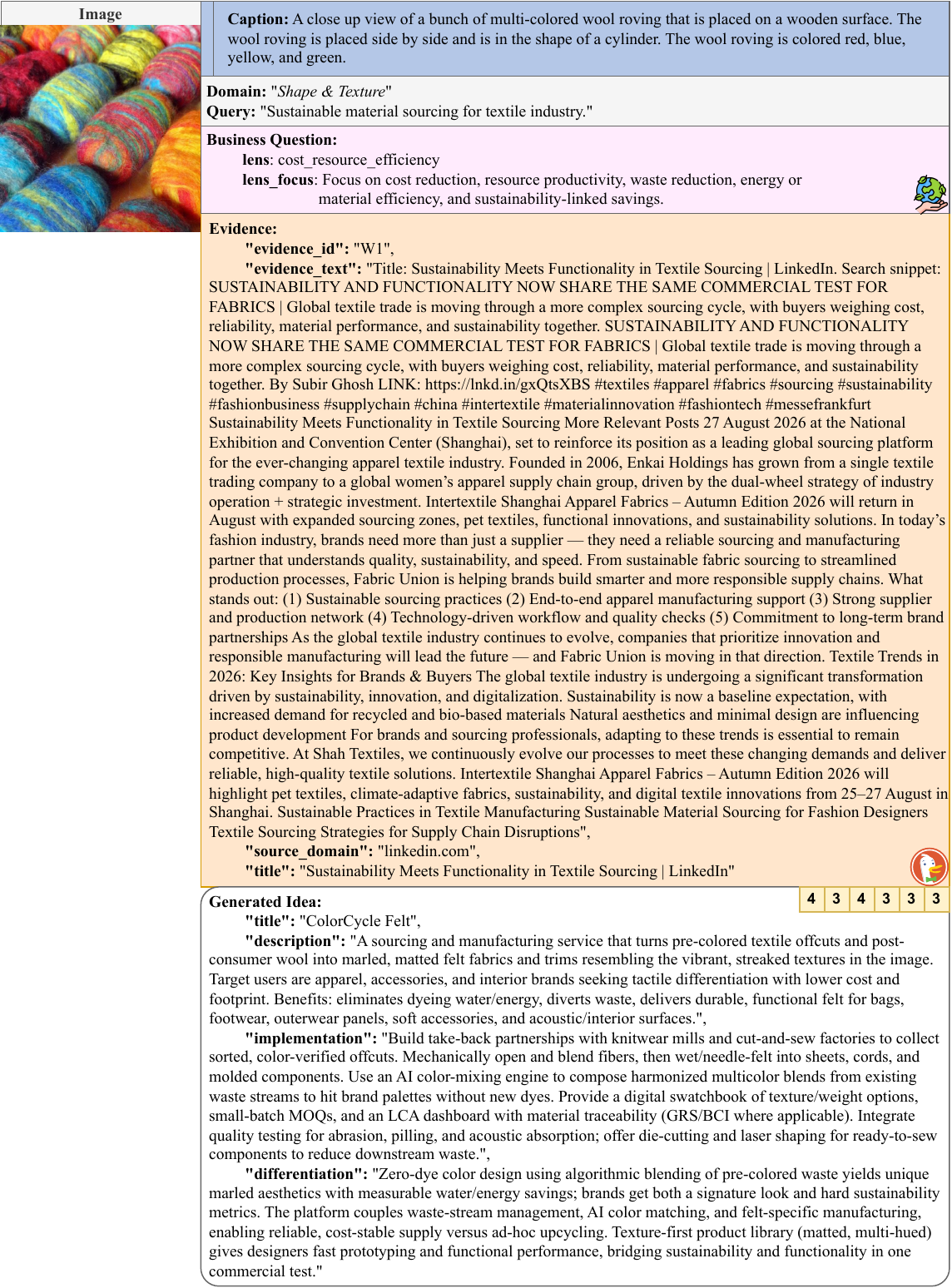}
  \caption{An example from the \textit{Shape \& Texture} domain in MBA-Bench.}
  \label{fig:mba_shape}
\end{figure*}

\begin{figure*}[!p]
  \centering
  \includegraphics[
    width=\textwidth,
    height=\textheight,
    keepaspectratio
  ]{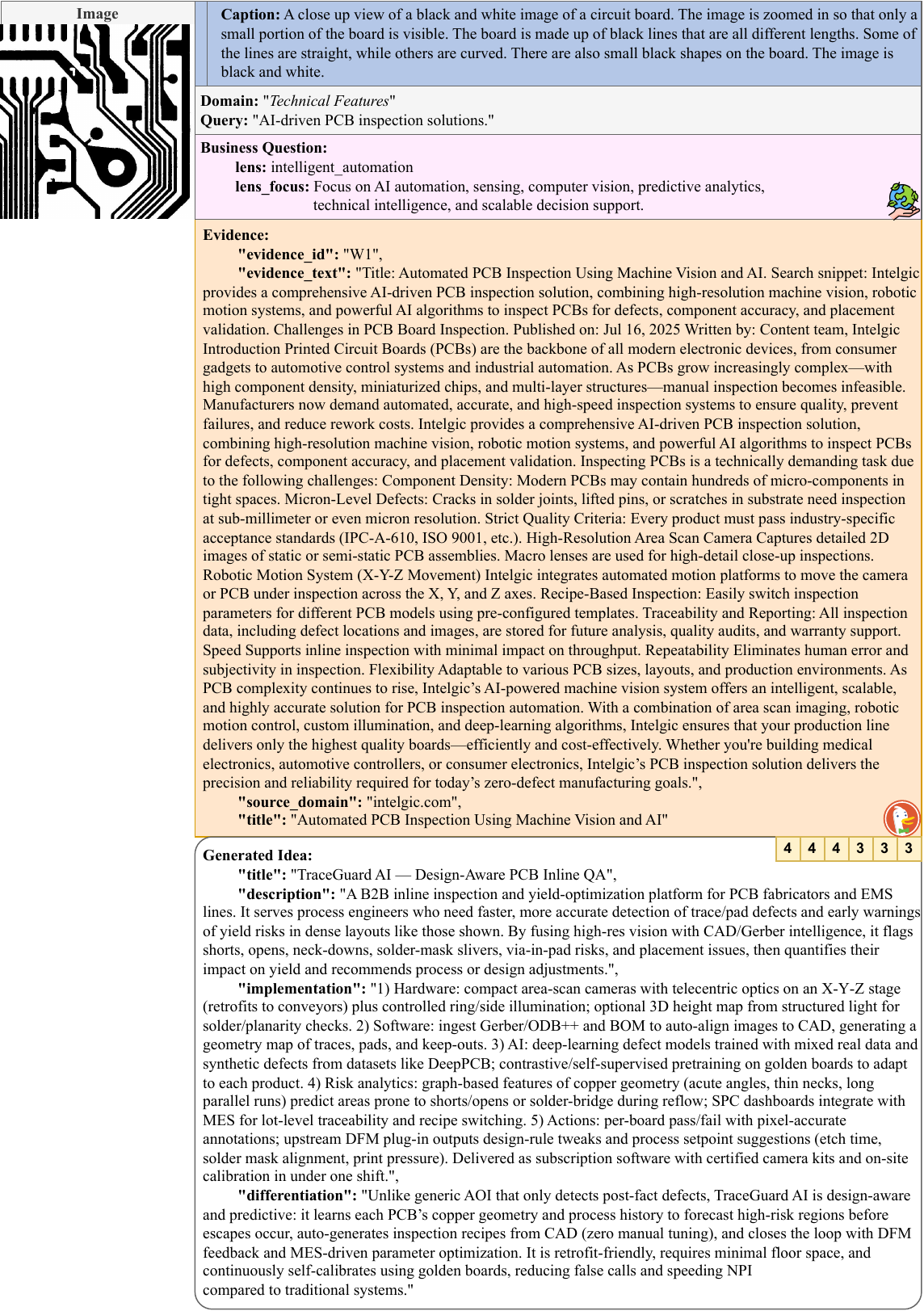}
  \caption{An example from the \textit{Technical Features} domain in MBA-Bench.}
  \label{fig:mba_tech}
\end{figure*}

Figures~\ref{fig:mba_general},~\ref{fig:mba_spatial},~\ref{fig:mba_crowd},~\ref{fig:mba_visual},~\ref{fig:mba_shape}, and~\ref{fig:mba_tech} present representative MBA-Bench samples from each visual domain. As in Figure~\ref{fig:f4}, the evaluation score for each metric is displayed in the upper-right corner of each generated idea box. Each example illustrates the complete benchmark instance, beginning with an image--caption pair and its corresponding domain, followed by the business question, generated query, and retrieved
evidence, and concluding with the business idea generated by our model. These examples demonstrate how MBA integrates visually grounded observations with task-specific questions and external knowledge to produce contextually relevant and actionable business ideas across diverse real-world settings.


\section{Prompts}

\subsubsection{Prompt for MBA-Bench}
\begin{figure*}[!p]
  \centering
  \includegraphics[
    width=\textwidth,
    height=\textheight,
    keepaspectratio
  ]{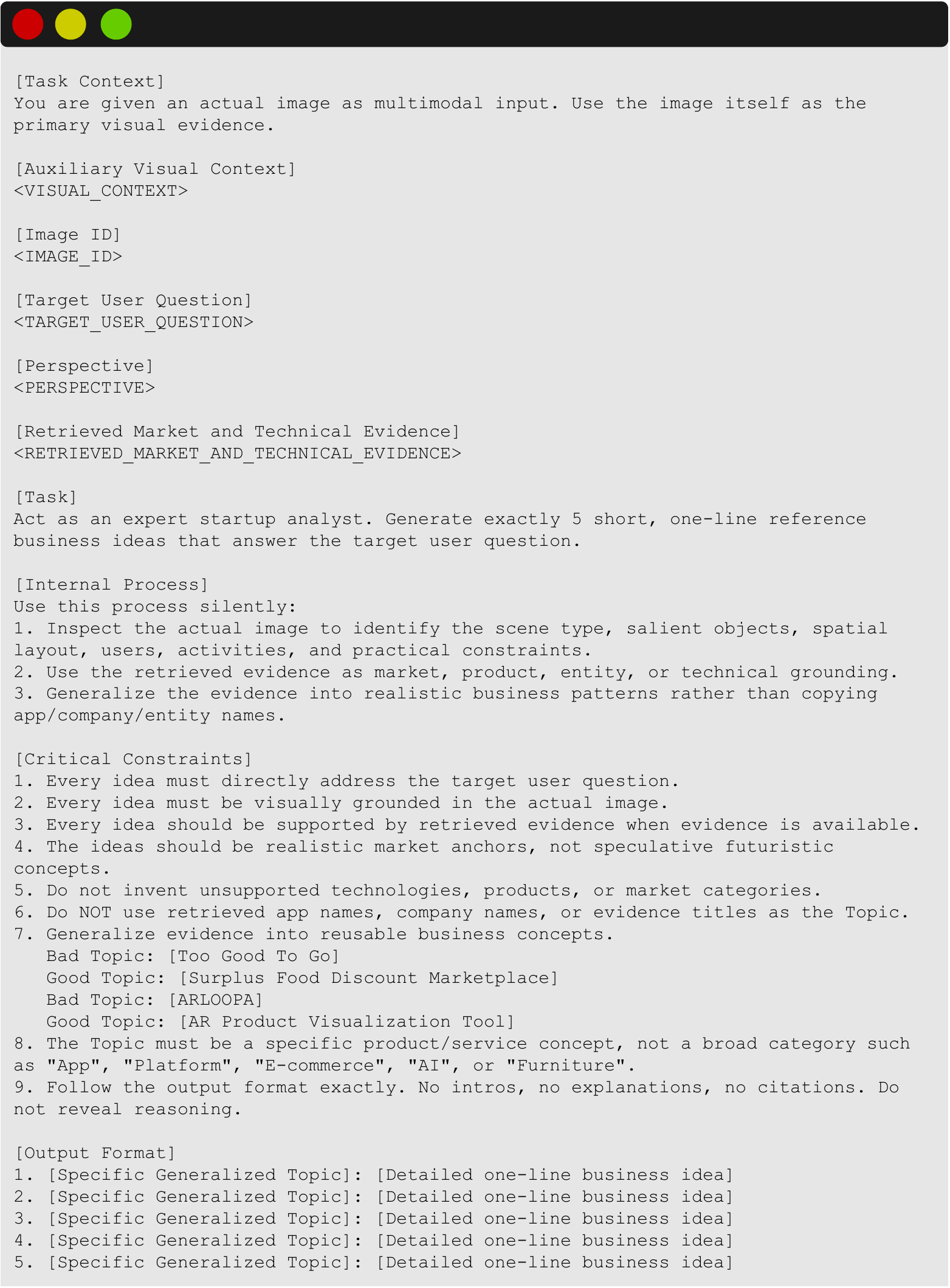}
  \caption{Prompt template used to construct MBA-Bench.}
  \label{fig:prompt-data}
\end{figure*}

The expert-data prompt in Figure~\ref{fig:prompt-data} provides the model with the input image, target user question, business perspective, and retrieved market and technical evidence. It instructs the model to produce five concise and visually grounded reference business ideas that serve as expert trajectories for the subsequent training stages.

\subsubsection{Prompt for SFT}
\begin{figure*}[!p]
  \centering
  \includegraphics[
    width=\textwidth,
    height=\textheight,
    keepaspectratio
  ]{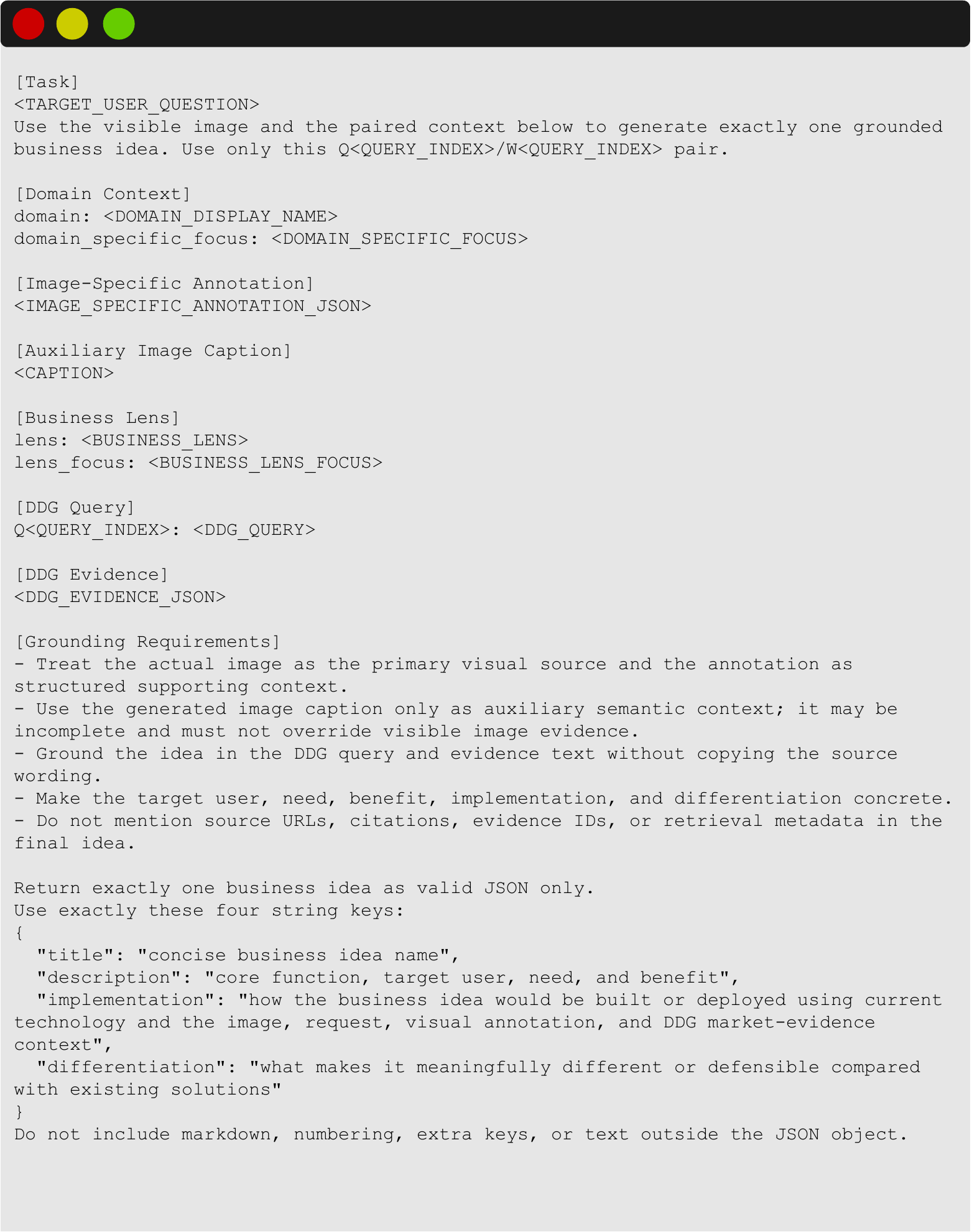}
  \caption{Prompt template used for supervised fine-tuning (SFT).}
  \label{fig:prompt-sft}
\end{figure*}

Figure~\ref{fig:prompt-sft} shows the prompt used for SFT. It combines the image with domain context, an auxiliary caption, image-specific annotations, a business lens, and paired market evidence, and requires the model to generate one grounded business idea using a four-field JSON format.

\subsubsection{Prompt for GRPO}
\begin{figure*}[!p]
  \centering
  \includegraphics[
    width=\textwidth,
    height=\textheight,
    keepaspectratio
  ]{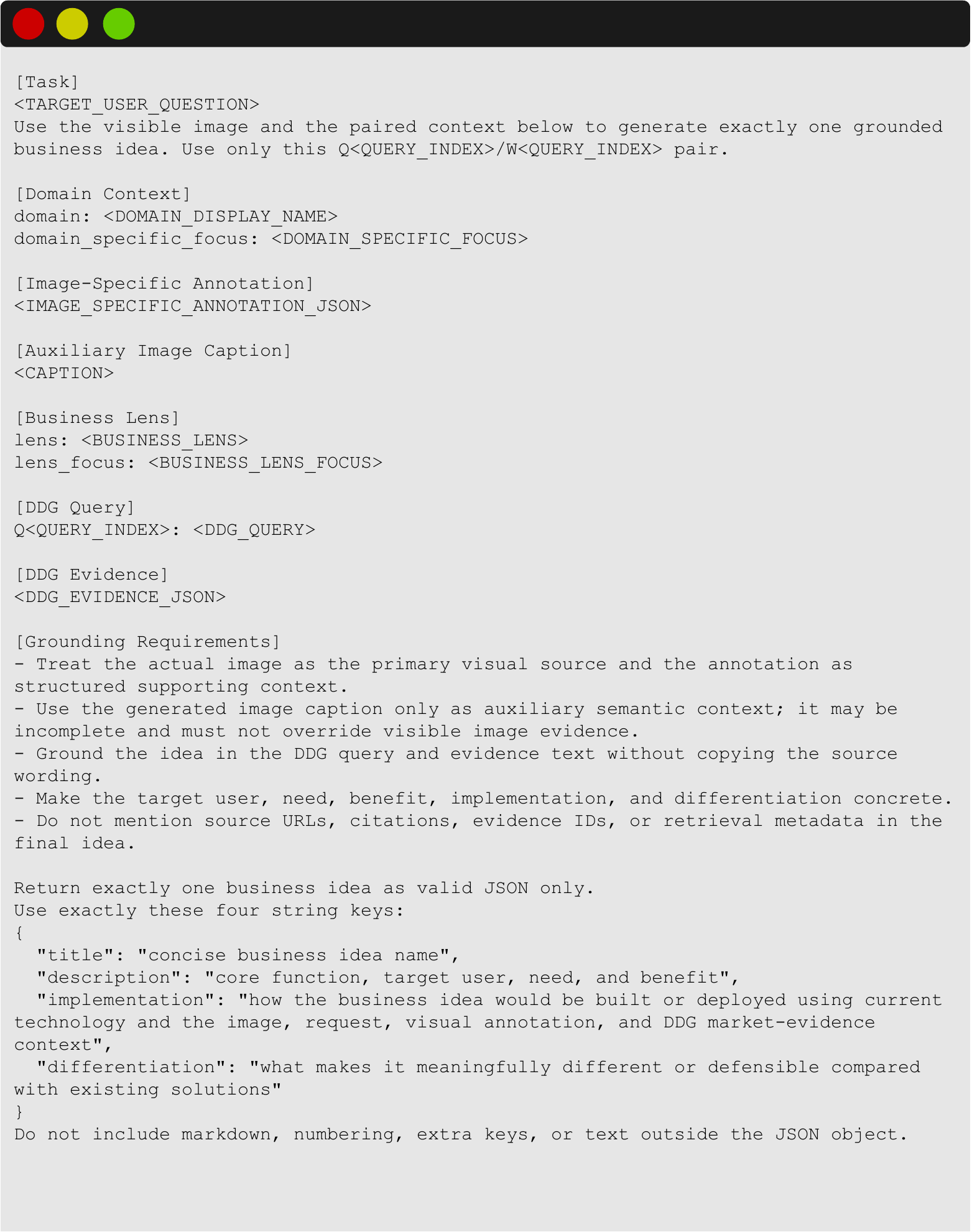}
  \caption{Prompt template used for GRPO policy rollouts.}
  \label{fig:prompt-grpo}
\end{figure*}

The GRPO prompt is presented in Figure~\ref{fig:prompt-grpo}. It uses the same canonical input structure as the SFT stage so that policy optimization changes the quality of the generated business ideas without introducing a different task formulation or output schema.

\subsubsection{Prompt for Evaluation}
\begin{figure*}[!p]
  \centering
  \includegraphics[
    width=\textwidth,
    height=\textheight,
    keepaspectratio
  ]{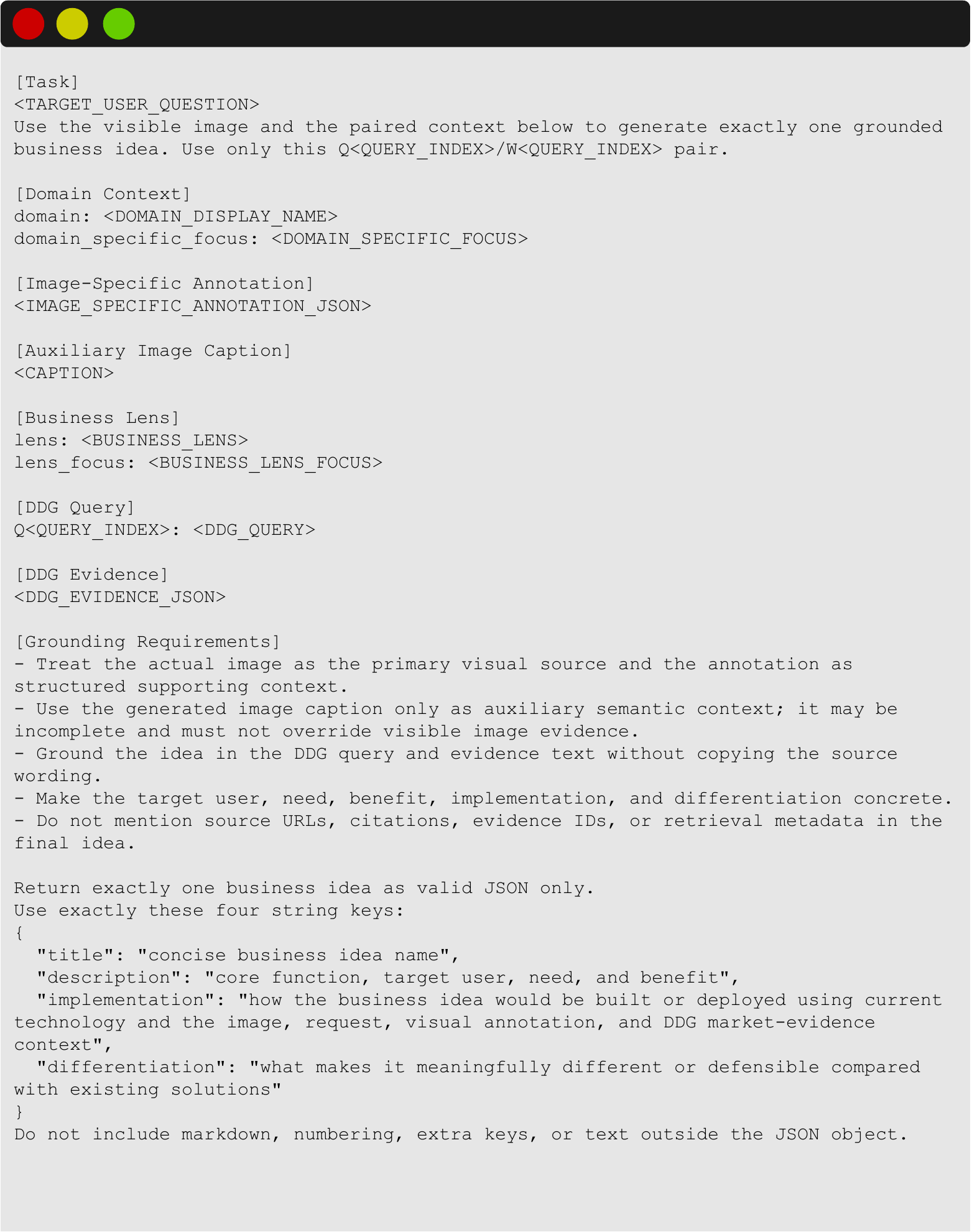}
  \caption{Prompt template used for evaluation.}
  \label{fig:prompt-eval}
\end{figure*}

Figure~\ref{fig:prompt-eval} presents the generation prompt used during evaluation. The same image-grounded context and structured JSON requirements are applied consistently across the evaluated models, enabling a controlled comparison under an identical generation setting.

\stopcontents[supplement]

\end{document}